\documentclass[10pt,a4paper]{article}
\usepackage[margin=1in]{geometry}

\usepackage[table]{xcolor} % This needs to appear before tikz.
\newcommand{\cellred}{\cellcolor{red!25}}
\newcommand{\cellgreen}{\cellcolor{green!25}}

\usepackage{graphicx}
\graphicspath{{figures/}}
\usepackage{tikz}

\usepackage{caption}
\usepackage{subcaption}

\usepackage{booktabs}
\usepackage{array}
\usepackage{xltabular}

\usepackage{amssymb,amsmath}

\usepackage{xurl}
\usepackage{hyperref}
\hypersetup{colorlinks=true,urlcolor=blue}

\usepackage[numbers,sort&compress]{natbib}

\begin{document}

\title{Deepfake: Definitions, Performance Metrics and Standards, Datasets and Benchmarks, and a Meta-Review}

\author{Enes Altuncu, Virginia N.~L.~Franqueira  
and Shujun Li\thanks{Corresponding author}\\
Institute of Cyber Security for Society (iCSS) \& School of Computing\\ University of Kent, UK\\
\{ea483, V.Franqueira, S.J.Li\}@kent.ac.uk}

\date{}
\maketitle 

\begin{abstract}
Recent advancements in AI, especially deep learning, have contributed to a significant increase in the creation of new realistic-looking synthetic media (video, image, and audio) and manipulation of existing media, which has led to the creation of the new term ``deepfake''. Based on both the research literature and resources in English and in Chinese, this paper gives a comprehensive overview of deepfake, covering multiple important aspects of this emerging concept, including 1) different definitions, 2) commonly used performance metrics and standards, and 3) deepfake-related datasets, challenges, competitions and benchmarks. In addition, the paper also reports a meta-review of 12 selected deepfake-related survey papers published in 2020 and 2021, focusing not only on the mentioned aspects, but also on the analysis of key challenges and recommendations. We believe that this paper is the most comprehensive review of deepfake in terms of aspects covered, and the first one covering both the English and Chinese literature and sources.
\end{abstract}


\section{Introduction\label{sec:intro}}

Recent advancements in AI and machine learning have increased the capability to produce more realistic media, e.g., video, image, and audio. Especially, state-of-the-art deep learning methods enabled the generation of ``deepfakes", manipulated or synthetic media the realness of which are not easily recognisable by the human eye. Although deepfake is a relatively new phenomenon (having first appeared at the end of 2017), its growth has been remarkable. According to the 2019 and 2020 Deeptrace reports on the state of deepfake~\cite{ajder2019}, the number of deepfake videos in the English-speaking internet grew from 7,964 (December 2018) to 14,678 (July 2019) to 85,047 (December 2020), representing a 968\% increase from 2018 to 2020.

In this work, we review existing deepfake-related research ecosystem in terms of various aspects, including performance metrics and standards, datasets, challenges, competitions, and benchmarks. Furthermore, we provide a meta-review of 12 selected deepfake-related survey papers which covers several additional aspects other than the mentioned ones in a systematic manner, such as performance comparison, key challenges, and recommendations.

Despite being a hugely popular term, there is a lack of consensus on the definition of ``deepfake'' and the boundary between deepfakes and non-deepfakes is not clear cut. For this survey, we adopt a relatively more inclusive approach to cover all forms of manipulated or synthetic media that are considered deepfakes in a broader sense. We also cover closely related topics including biometrics and multimedia forensics, since deepfakes are often used to launch presentation attacks against biometrics-based authentication systems and detection of deepfakes can be considered part of multimedia forensics. A more detailed discussion on different definitions of ``deepfake'' is given next.

\subsection{Definitions of the Term Deepfake\label{sec:definitions}}

As its name implies, the term ``deepfake'' is derived from the combination of ``deep'' (referring to \emph{deep learning} (DL)) and ``fake''. It is normally used to refer to manipulation of existing media (image, video and/or audio) or generation of new (synthetic) media using DL-based approaches. The most commonly discussed deepfake data are fake face images, fake speech forgeries, and fake videos that combine both fake images and fake speech forgeries. While having ``fake'' in the word indicates manipulated or synthesised media, there are plenty of benign applications of the deepfake technology, e.g., for entertainment and creative arts. With this respect, another term ``deep synthesis'' has been proposed as a more neutral-sounding alternative~\cite{tencent2020}. This new term, however, has not been widely adopted.

In addition to the lack of a universal definition, as mentioned already, the boundary between deepfakes and non-deep fakes is actually not a clear cut. There are at least two important aspects we should consider, one on detection of and the other on creation of deepfakes.

First, detection of deepfakes often follows very similar approaches to detection of traditional fakes generated without using DL techniques. Advanced detection methods have also started leveraging DL to improve their performance, but they do not necessarily need to know how a target media is created (deep or not). To some extent, one could argue that detecting deepfakes does not involve developing deepfake-specific methods (even though some researchers choose to do so), but a more robust and universal detector that can handle any (deep or not) fake media. This can be seen for two closely related topics: biometrics and multimedia forensics. For biometrics, there is a trend of using deep learning techniques to generate fake biometric signals (e.g., face images and videos) for biometric spoofing or presentation attacks. For multimedia forensics, deepfake-based forgeries have become a new threat to the traditional problem of ``forgery detection''. For both topics, detection of biometric spoofing and multimedia forgeries have evolved to consider both deep and non-deep fakes.

Second, one may argue that the word ``deep'' in ``deepfake'' does not necessarily refer to the use of ``deep learning'', but any ``deep'' (i.e., sophisticated) technology that creates a very believable fake media. For instance, \citet{Brady2020} considered deepfake as audio-visual manipulation using ``a spectrum of technical sophistication ... and techniques''. They also introduced two new terms, \emph{Shallowfake} and \emph{Cheapfake}, referring to ``low level manipulation of audio-visual media created with (easily) accessible software [or no software] to speed, slow, restage or re-contextualise content''. This broader understanding of ``deepfake'' has also been adopted by law makers for new legislations combating malicious deepfakes. For instance, the following two United States acts define ``deepfakes'' as follows:
\begin{itemize}
\item 2018 Malicious Deep Fake Prohibition Act\footnote{\url{https://www.congress.gov/bill/115th-congress/senate-bill/3805}}:\\
§1041.(b).(2): ``\textit{the term `deep fake' means an audiovisual record created or altered in a manner that the record would falsely appear to a reasonable observer to be an authentic record of the actual speech or conduct of an individual.}''

\item 2019 DEEP FAKES Accountability Act\footnote{\url{https://www.congress.gov/bill/116th-congress/house-bill/3230}}:\\
§1041.(n).(3): ``{\itshape The term `deep fake' means any video recording, motion-picture film, sound recording, electronic image, or photograph, or any technological representation of speech or conduct substantially derivative thereof—\\
(A) which appears to authentically depict any speech or conduct of a person who did not in fact engage in such speech or conduct; and\\
(B) the production of which was substantially dependent upon technical means, rather than the ability of another person to physically or verbally impersonate such person.}''
\end{itemize}
As we can see from the above legal definitions of ``deepfake'', the use of DL as a technology is not mentioned at all. The focus here is on ``authenticity'', ``impersonation'' and (any) ``technical means''.

\subsection{Scope and Contribution\label{Subsec:contribution}}

Based on the above discussion on definitions of deepfake, we can see it is not always straightforward or meaningful to differentiate deepfakes from non-deep fakes. In addition, for our focus on performance evaluation and comparison, the boundary between deepfakes and non-deep fakes is even more blurred. This is because DL is just a special (deeper) form of machine learning (ML), and as a result, DL and non-deep ML methods share many common concepts, metrics and procedures.

Despite the fact that deepfake may be understood in a much broader sense, in this work, we have a sufficiently narrower focus to avoid covering too many topics. We, therefore, decided to define the scope of this survey as follows:

\begin{itemize}
\item For metrics and standards, we chose to include all commonly used ones for evaluating general ML methods and those specifically defined for evaluating deepfake creation or detection methods.

\item For datasets, challenges, competitions and benchmarks, we considered those related to fake media covered in the deepfake-related survey papers and those with an explicit mention of the term ``deepfake'' or a comparable term.

\item For the meta-review, we considered only survey papers whose authors explicitly referred to the term ``deepfakes'' in the meta data (title, abstract and keywords).
\end{itemize}

\section{Methodology\label{sec:methodology}}

Research papers covered in this survey (i.e., the deepfake-related survey papers) were identified via systematic searches on the scientific databases, Scopus and China Online Journals (COJ)\footnote{\url{https://c.wanfangdata.com.cn/periodical}}. The following search queries were used to perform the searches on Scopus and COJ, respectively:

\begin{quote}
(deepfake* OR deep-fake* OR ``deep fake*") AND (review OR survey OR overview OR systemati* OR SoK)
\end{quote}

\begin{quote}
%(deepfake OR 深度伪造) AND (综述 OR 进展)
\includegraphics[height=0.94\baselineskip]{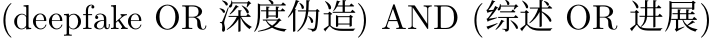}
\end{quote}

The searches returned 41 survey papers in English and 15 survey papers in Chinese. Out of these papers, eight published in English and four published in Chinese were selected for consideration. 

Deepfake-related challenges, competitions and benchmarks were identified via multiple sources: the survey papers selected, research papers from the co-authors' personal collections, Google Web searches, and manual inspection of websites of major AI-related conferences held in 2020 and 2021 (where such challenges and competitions are routinely organised). The inspected conferences include those listed in the ACL (Association for Computational Linguistics) Anthology\footnote{\url{https://aclanthology.org/}}, ICCV, CVPR, AAAI, ICML, ICLR, KDD, SIGIR, WWW, and many others. In addition, a comprehensive list of datasets was compiled based on the selected survey papers and the identified challenges, competitions, and benchmarks. Relevant standards were identified mainly via research papers covered in this survey, the co-authors' personal knowledge, and Google Web searches. For performance metrics, we covered those commonly used based on relevant standards, the survey papers, and the identified challenges, competitions, and benchmarks.

\section{Deepfake-Related Performance Metrics \& Standards\label{sec:metrics-and-standands}}

In this survey, we focus on performance evaluation and comparison of deepfake generation and detection methods. The metrics used for such performance evaluations are at the core of our discussions. In this section, we review the performance metrics that are commonly used to evaluate deepfake generation and detection algorithms. Note that all metrics covered in this section are also commonly used for evaluating performance of similar systems that are not for generating or detecting deepfakes. Therefore, this section can be seen as a very brief tutorial on general performance metrics.

In the last subsection, we also briefly discuss how the related performance metrics are covered in formal standards. By ``formal standards'', we refer to standards defined following a formal procedure, often by one or more established standardisation bodies such as the International Organization for Standardization (ISO)\footnote{\url{https://www.iso.org/}} and the International Electrotechnical Commission (IEC)\footnote{\url{https://www.iec.ch/}}. Note that we consider a broad range of documents defined to be standards by standardisation bodies, e.g., International Telecommunication Union (ITU)\footnote{\url{https://www.itu.int/}} recommendations and ISO technical reports (TRs).

\subsection{The Confusion Matrix}
\label{Subsec:confusion-matrix}

Deepfake detection is primarily a binary classification problem. A binary classifier takes an input that is \textit{actually positive} or \textit{actually negative} and outputs a binary value denoting it to be \textit{predicted positive} or \textit{predicted negative}. For example, a deepfake detection system will take a suspected image as the input that may be \textit{actually fake} or \textit{actually real} and output \textit{predicted fake} or \textit{predicted real}.

A fundamental tool used in evaluating a binary classifier is the \textbf{confusion matrix} that summarises the success and failure of the classification model. On one axis are the two \textit{actual} values and on the other axis are the two \textit{predicted} values. The classification is \textit{successful/correct/true} (true positive and true negative) when the actual and the predicted values match. It is \textit{failed/incorrect/false} (false positive and false negative) when the actual and predicted values do not match. Table~\ref{tab:confusion-matrix-for-fake-detection} shows the confusion matrix for a binary deepfake classifier (detector). The two cells in green, TP (the number of \textbf{true positives}) and TN (the number of \textbf{true negatives}), indicate correct prediction results, and the two cells in red, FN (the number of \textbf{false negatives}) and FP (the number of \textbf{false positives}), indicate two different types of errors when making incorrect prediction results.

\begin{table}[!h]
\centering
\caption{Confusion matrix for a binary classifier for detecting deepfake.}
\label{tab:confusion-matrix-for-fake-detection}
\begin{tabular}{|c|c|c|}
\hline
 & fake (predicted) & real (predicted)\\
\hline
fake (actual) & \cellgreen TP & \cellred FN\\
\hline
real (actual) & \cellred FP & \cellgreen TN\\
\hline
\end{tabular}
\end{table}

\subsection{Precision and Recall}

Based on the four fundamental values introduced in Section~\ref{Subsec:confusion-matrix}, i.e., TP, TN, FP and FN, we define two important performance metrics for a binary classifier -- \textbf{precision} and \textbf{recall}.

Precision of a binary classifier is defined as the fraction of \textit{actually positive} samples among all the \textit{predicted positives}. In the confusion matrix, it is the fraction of true samples in the first column. It can be formally defined as Eq.~\eqref{eq:precision}.

\begin{align}
\label{eq:precision}
\text{precision} = \frac{\text{TP}}{\text{TP} + \text{FP}}
\end{align}

When the ``natural'' ratio between positive and negative samples is significantly different from the test set, it is often useful to adjust the weight of the false positives, which leads to the \textbf{weighted precision} (wP) defined in Eq.~\eqref{eq:WP}, where $\alpha>0$ is a weight determined by the ratio between the negative and positive samples.

\begin{align}
\text{wP} & = \frac{\text{TP}}{\text{TP} + \alpha\text{FP}} \label{eq:WP}
\end{align}

Recall of a binary classifier is the fraction of \textit{predicted positive} samples among the \textit{actually positive} samples, as shown in Eq.~\eqref{eq:recall}. In the confusion matrix, it is the fraction of true samples in the first row.
\begin{align}
\label{eq:recall}
\text{recall} = \frac{\text{TP}}{\text{TP} + \text{FN}}
\end{align}

Let us consider an example binary classifier that predicts  if an image from a database containing both deepfake and real (authentic) images is fake or not. Precision of the classifier is the fraction of correctly classified images among all images classified as deepfake. On the other hand, recall is the fraction of deepfake images identified by the classifier, among all deepfake images in the database.

\subsection{True and False Positive Rates}

Focusing on predicted positive samples, we can also define two metrics: \textbf{true positive rate} (TPR), also called \textbf{correct detection rate} (CDR), as the fraction of the predicted positive samples among the actually positive samples and \textbf{false positive rate} (FPR), also called \textbf{false alarm rate} (FAR), as the fraction of the predicted positive samples among the actually negative samples, as shown in Eqs.~\eqref{eq:TPR} and \eqref{eq:FPR}. In the confusion matrix, TPR is the fraction of predicted positive samples in the first row and FPR is the fraction of predicted positive samples in the second row. Note that TPR is basically a different name for \textbf{recall} (Eq.~\eqref{eq:recall}). 

\begin{align}
\label{eq:TPR}
\text{TPR} = \frac{\text{TP}}{\text{TP} + \text{FN}}
\end{align}
\begin{align}
\label{eq:FPR}
\text{FPR} = \frac{\text{FP}}{\text{FP} + \text{TN}}
\end{align}

\subsection{True and False Negative Rates}

Similar to true and false positive rates, we can define two other rates focusing on negative predicted results: \textbf{true negative rate} (TNR) indicating the fraction of the predicted negative samples among the actually negative samples, and \textbf{false negative rate} (FNR) indicating the fraction of the predicted negative samples among the actually positive samples, as shown in Eqs.~\eqref{eq:TNR} and \eqref{eq:FNR}.
\begin{align}
\label{eq:TNR}
\text{TNR} = \frac{\text{TN}}{\text{TN} + \text{FP}}
\end{align}
\begin{align}
\label{eq:FNR}
\text{FNR} = \frac{\text{FN}}{\text{FN} + \text{TP}}
\end{align}

\subsection{Sensitivity and Specificity}

In some applications of binary classifiers, especially in biology and medicine, the TPR and the TNR are more commonly used, and they are often called \textbf{sensitivity} (TPR) and \textbf{specificity} (TNR). The focus of these two terms is on the two types of correctness of the predicted results. These are less used in deepfake-related research, hence, we will not refer to them in the remainder of this paper.

\subsection{Equal Error Rate}

Focusing on error rates means that we need to consider the FPR and the FNR. These two rates normally conflict with each other so that reducing one rate normally leads to an increase in the other. Therefore, rather than trying to reduce both error rates at the same time, which is normally impossible, the more realistic task in practical applications is to find the right balance so that they are both below an acceptable threshold.

In some applications, such as biometrics, people are particularly interested in establishing the so-called \textbf{equal error rate} (EER) or \textbf{crossover error rate} (CER), the point where the FPR and the FNR are equal. The EER/CER is not necessarily a good metric for some applications, especially when the two types of errors are of different levels of importance, e.g., for detecting critical deepfakes (e.g., fake news that can influence how people cast their votes) we can often tolerate more false positives (false alarms) than false negatives (missed alarms).

\subsection{Accuracy and F-Score}

In addition to the EER/CER, there are also other metrics that try to reflect both types of errors, in order to give a more balanced indication of the overall performance of a binary classifier. The two most commonly used are \textbf{accuracy} and \textbf{F-score} (also called \textbf{F-measure}). Both metrics can be defined based on the four fundamental values (TP, TN, FP, and FN).

Accuracy of a binary classifier is defined as the fraction of \textit{correctly predicted} samples (true positives and true negatives) among the total number of samples that have been classified, as shown in Eq.~\eqref{eq:accuracy}.

\begin{align}
\label{eq:accuracy}
\text{accuracy} = \frac{\text{TP} + \text{TN}}{\text{TP} + \text{TN} + \text{FP} + \text{FN}}
\end{align}

The F-score of a binary classifier is actually a family of metrics. Its general form can be described based on a parameter $\beta$ as defined in Eq.~\eqref{eq:F-beta}.

\begin{equation}
F_{\beta} = (1+\beta^2)\cdot\frac{\text{precision}\cdot\text{recall}}{\beta^2\cdot\text{precision}+\text{recall}} \label{eq:F-beta}
\end{equation}

The most widely used edition of all F-scores is the so-called \textbf{F1-score}, which is effectively the F-score with $\beta=1$. More precisely, it is defined as shown in Eq.~\eqref{eq:F1}.

\begin{equation}
F_1 = 2\cdot\frac{\text{precision}\cdot\text{recall}}{\text{precision}+\text{recall}} = \frac{2\text{TP}}{2\text{TP}+\text{FP}+\text{FN}} \label{eq:F1}
\end{equation}
\vspace{0.1cm}

\subsection{Receiver Operating Characteristic Curve and Area Under Curve}

\textbf{Receiver operating characteristic} (ROC) curves are commonly used to measure the performance of binary classifiers that output a score (or probability) of prediction. 

Consider the following. Let $S$ be the set of all test samples and let the output scores $f(s)$ (for all $s \in S$) lie in the interval $[a, b]$ on the real line.
Let $t \in [a,b]$ be a prediction threshold for the model, and assume that the classifiers works as follows for all $s \in S$:
\begin{equation}
\text{class}(s) =
\left\{
\begin{array}{ll}
    \text{positive}, & \text{if } f(s) \geq t, \text{ and}\\
    \text{negative}, & \text{otherwise}.
\end{array}
\right.
\end{equation}

It is easy to see that, for $t = a$, all the samples will be classified as positive, leading to $\text{FN}=\text{TN}=0$ so $\text{TPR}=\text{FPR}=1$; while for $t = b$, all the samples will be classified as negative, leading to $\text{FP}=\text{TP}=0$ so $\text{TPR}=\text{FPR}=0$. For other threshold values between $a$ and $b$, the values of TPR and FPR will normally be between 0 and 1. By changing $t$ from $a$ to $b$ continuously, we can normally get a continuous curve that describes how the TPR and FPR values change from (0,0) to (1,1) on the 2D plane. This curve is the ROC curve of the binary classifier. 

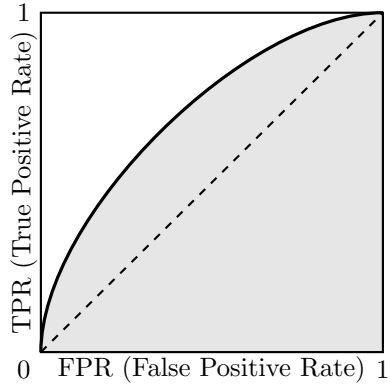
\begin{figure}[!h]
\centering
\begin{tikzpicture}[scale=0.45]
\draw[fill=black!10] (0,0) .. controls (0,4) and (6,10) .. (10,10) -- (10,0) -- (0,0);
\draw[thick] (0,0) rectangle (10,10);
\node[rotate=90] at (-0.5, 5) {TPR (True Positive Rate)};
\node[rotate=0] at (-0.5, -0.5) {0};
\node[rotate=0] at (5, -0.5) {FPR (False Positive Rate)};
\node[rotate=0] at (-0.5, 10) {1};
\node[rotate=0] at (10, -0.5) {1};
\draw[dashed, thick] (0,0) -- (10,10);
\draw[very thick] (0,0) .. controls (0,4) and (6,10) .. (10,10);
\end{tikzpicture}
\caption{A representative ROC curve showing how TPR and FPR change w.r.t.\ the (hidden) threshold $t$. The area under the (ROC) curve (AUC) is shown in grey.}
\label{fig:area-under-ROC-curve-example}
\end{figure}

For a random classifier, assuming that $f(s)$ distributes uniformly on $[a, b]$ for the test set, we can mathematically derive its ROC curve being the $\text{TPR}=\text{FPR}$ line, whose area under the ROC curve (AUC) is 0.5. For a binary classifier that performs better than a random predictor, we can also mathematically prove that its AUC is always higher than 0.5, with 1 being the best possible value. Note that no binary classifier can have an AUC below 0.5, since one can simply flip the prediction result to get a better predictor with an AUC of $1-\text{AUC}$. The relationship between the ROC and the AUC is graphically illustrated in Figure~\ref{fig:area-under-ROC-curve-example}.

\subsection{Log Loss}

Another widely used performance metric for binary classifiers that can return a probability score for the predicted label is \textbf{log loss}. For a binary classification with a true label $y\in\{0,1\}$ and an estimated probability  $p=\text{Pr}(y=1)$, the log loss per sample is the negative log-likelihood of the classifier given the true label, defined as shown in Eq.~\eqref{eq:Llog}.

\begin{equation}
L_{\log}(y,p)=-(y\log(p)+(1-y)\log(1-p)) \label{eq:Llog}
%Source: https://scikit-learn.org/stable/modules/model_evaluation.html#log-loss
\end{equation}

Given a testing set with $n$ samples, the log loss score of a binary classifier can be calculated using Eq.~\eqref{eq:LL}, where $y_i$ is 1 if the $i$-th sample is true and 0 if false, and $\hat{y}_i$ is the predicted probability of $y_i=1$.

\begin{equation} \label{eq:LL}
\text{LL}=-\frac{1}{n}\sum_{i=1}^n[y_i\log(\hat{y}_i)+(1-y_i)\log(1-\hat{y}_i)]
%Source: https://www.kaggle.com/c/deepfake-detection-challenge/overview/evaluation
\end{equation}

\subsection{Extension to Multi-class Classifiers}

All metrics that are defined based on the four basic values TP, TN, FP and FN can be easily extended to \textbf{multi-class classification} by considering the prediction to be true or false individually with respect to each class. For example, if the system is classifying animals (cats, dogs, horses, lions, tigers, etc.), then a true positive prediction of an image to be of a cat, would simultaneously be true negative predictions for the remaining classes (dogs, horses, lions, tigers, etc.). If an image of a cat is incorrectly predicted to be that of a dog, it would be a false negative with respect to a cat, a false positive with respect to a dog, and a true negative with respect to all other classes.

\subsection{Perceptual Quality Assessment (PQA) Metrics}

By definition, the main goal of deepfakes is to make it hard or impossible for human consumers (listeners or viewers) to distinguish fake media from real media. Therefore, when evaluating the quality of deepfake media, the quality perceived by human consumers of the media is key. This calls for subjective assessment of the perceptual quality of the deepfake media as the ``gold standard''. The most widely used subjective perceptual quality assessment (PQA) metric for audio-visual signals is \textbf{mean opinion score} (MOS), which has been widely used by the signal processing and multimedia communication communities, including digital TV and other multimedia-related consumer applications. As its name implies, MOS is calculated by averaging the subjective scores given by a number of human judges, normally following a numerical scale between 1 and 5 or between 0 and 100. MOS has been used in some deepfake-related challenges (see Section~\ref{Subsec:challenges_deepfake_generation}) and also for evaluating and comparing the quality (realness/naturalness) of deepfake datasets (see Section~\ref{Subsec:datasets_subjective_quality}).

As a general subjective PQA metric, MOS has been standardised by the ITU\footnote{\url{https://www.itu.int/rec/T-REC-P.800.1-201607-I/en}}. There are also ITU standards defining more specific subjective Video Quality Assessment (VQA) metrics and the standard procedures one should follow to conduct VQA user studies, e.g., ITU-T Recommendation P.910 ``Subjective video quality assessment methods for multimedia applications''\footnote{\url{https://www.itu.int/rec/T-REC-P.910-200804-I/en}}. Note that the ITU standards focus more on traditional perceptual quality, i.e., how good a signal looks or sounds, even if it looks or sounds not real (e.g., too smooth). On the other hand, for deepfakes, the focus is rather different because what matters is the realness and naturalness of the created media, i.e., how real and natural it looks or sounds, even if it is of low quality. To some extent, we can also consider realness and naturalness as a special aspect of perceptual quality.

One major problem of subjective PQA metrics like MOS is the need to recruit human judges and to have a well-controlled physical testing environment and protocol, which are not easy for many applications. To help reduce the efforts and costs of conducting PQA-related user studies, various objective PQA metrics have been proposed, where the term ``objective'' refers to the fact that such metrics are human-free, i.e., automatically calculated following a computational algorithm or process. Depending on whether a reference exists, such objective PQA metrics can be largely split into three categories: full-reference (FR) metrics (when the original ``perfect-quality'' signal is available as the reference), reduced-reference (RR) metrics (when some features of the original ``perfect-quality'' signal are available as the reference), and no-reference (NR) metrics (when the original signal is unavailable or such an original signal does not exist). For deepfakes, normally NR or RR metrics are more meaningful because the ``fake'' part of the word means that part of the whole data does not exist in the real world, hence a full reference cannot be obtained. RR metrics are still relevant because deepfakes are often produced for a target's specific attributes (e.g., face and voice), where the reduced reference will be such attributes. NR metrics will be useful to estimate the realness and naturalness of a deepfake, simulating how a human judge would rate it in a controlled subjective PQA user study.

PQA is a very active research area and many PQA metrics have been proposed, some of which have been widely used in real-world products and services, e.g., \textbf{mean squared error} (MSE), \textbf{peak signal-to-noise ratio} (PSNR) and \textbf{structural similarity index measure} (SSIM) for FR PQA of digital images and videos defined as in Eqs.~\eqref{eq:MSE},~\eqref{eq:PSNR}, and~\eqref{eq:SSIM}, respectively, where $X=\{x_i\}_{i}^n$ is the reference (the original signal), $Y=\{y_i\}_{i}^n$ is the signal whose visual quality is assessed, $n$ is the number of pixels in $X$ and $Y$, $L$ is the maximum possible pixel value of $X$ and $Y$ (e.g., 255 for 8-bit gray-scale images), $c_1=(k_1L)^2$ and $c_2=(k_2L)^2)$ are two stabilising parameters ($k_1=0.01$ and $k_2=0.03$ by default). For more about PQA metrics for different types of multimedia signals, we refer readers to some relevant surveys \cite{ZX2020, AF2017, PR2018}.

\begin{equation}
\text{MSE}(X,Y) = \sum_{i=1}^n (y_i-x_i) \label{eq:MSE}
\end{equation}
\begin{equation}
\text{PSNR}(X,Y) = 10\log_{10}\left(\frac{L^2}{\text{MSE}}\right) \label{eq:PSNR}
\end{equation}
\begin{equation}
\text{SSIM}(X,Y) = \frac{(2\mu_x\mu_y+c_1)(2\sigma_{xy}+c_2)}{(\mu_x^2+\mu_y^2+c_1)(\sigma_x^2+\sigma_y^2+c_2)} \label{eq:SSIM}
\end{equation}

\subsection{More about Standards}

Many of the basic performance metrics described in this section have been widely used by deepfake researchers as de facto standards, e.g., EER, log loss and MOS have been widely used in deepfake-related challenges (see Section~\ref{Sec:challenges}). Also, the combination of precision, recall and F1-score has been widely used to assess performance of binary classifiers. While there have been a number of ITU standards on PQA to date, there does not seem to be many standardisation efforts on the performance metrics for evaluation of binary classifiers. This was the case until at least 2017, when ISO and IEC jointly set up the ISO/IEC JTC 1/SC 42\footnote{\url{https://www.iso.org/committee/6794475.html}}, a standardisation subcommittee (SC) focusing on AI under ISO/IEC JTC 1\footnote{\url{http://www.iso.org/iso/jtc1_home.html}}, the joint technical committee for standardising ``information technology''.

One recent effort that ISO/IEC JTC 1/SC 42 made is to produce the ISO/IEC TR 24029-1:2021 ``Artificial Intelligence (AI) -- Assessment of the robustness of neural networks -- Part 1: Overview''\footnote{\url{https://www.iso.org/standard/77609.html}}, a technical report (TR) that systematically covers many commonly used performance assessment concepts, methods and metrics. Although the technical report has ``neural networks'' in its title, most performance assessment concepts, methods and metrics included are common ones for all supervised machine learning models.

In terms of performance metrics, two other ongoing work items of the ISO/IEC JTC 1/SC 42 that deserve attention are as follows:
\begin{itemize}
\item ISO/IEC DTS (Draft Technical Specification) 4213 ``Information technology -- Artificial Intelligence -- Assessment of machine learning classification performance''\footnote{\url{https://www.iso.org/standard/79799.html}}

\item ISO/IEC AWI (Approved Work Item) TS (Technical Specifications) 5471 ``Artificial intelligence -- Quality evaluation guidelines for AI systems''\footnote{\url{https://www.iso.org/standard/82570.html}}
\end{itemize}

While the ISO/IEC JTC 1/SC 42 was created very recently, another standardisation subcommittee under ISO/IEC JTC1 has a much longer history of nearly 20 years: the ISO/IEC JTC 1/SC 37\footnote{\url{https://www.iso.org/committee/313770.html}} that focuses on biometrics-related technology. This standardisation subcommittee is highly relevant for deepfake since deepfake faces can be used to spoof biometrics-based user authentication systems. In this context, the following three standards are of particular relevance:

\textbf{ISO/IEC 19795-1:2021 ``Information technology -- Biometric performance testing and reporting -- Part 1: Principles and framework''\footnote{\url{https://www.iso.org/standard/73515.html}}}: This standard covers general metrics about evaluating biometric systems. Two major metrics in this context are \textbf{false accept rate} (FAR) and \textbf{false reject rate} (FRR), which refer to the standard FPR and FNR, respectively. This standard also deprecates the use of single-number metrics including the EER and AUC (which were widely used in biometrics-related research in the past).

\textbf{ISO/IEC 30107-1:2016 ``Information technology -- Biometric presentation attack detection -- Part 1: Framework''\footnote{\url{https://www.iso.org/standard/53227.html}}}: This standard defines a general framework about \textbf{presentation attack detection} (PAD) mechanisms, where the term ``\textbf{presentation attack}'' refers to the ``\textit{presentation of an artefact or of human characteristics to a biometric capture subsystem in a
fashion intended to interfere with system policy}''. It focuses on biometric recognition systems, where a PAD mechanism is a binary classifier trying to predict presentation attacks (also called attack presentations, e.g., fake faces) as positive and bona fide (real) presentations as negative.

\textbf{ISO/IEC 30107-3:2017 ``Information technology -- Biometric presentation attack detection -- Part 3: Testing and reporting''\footnote{\url{https://www.iso.org/standard/67381.html}}}: This standard defines a number of special performance metrics for evaluating PAD mechanisms standardised in the ISO/IEC 30107-1:2016. Three such metrics look at error rates: \textbf{attack presentation classification error rate} (APCER) referring to the standard FPR, \textbf{normal/bona fide presentation classification error rate} (NPCER/BPCER) referring to the standard FNR, and \textbf{average classification error rate} (ACER) that is defined as the average of the APCER and the NPCER/BPCER. Such metrics have been used in biometrics-related challenges such as Face Anti-spoofing (Presentation Attack Detection) Challenges\footnote{\url{https://sites.google.com/qq.com/face-anti-spoofing/}}. When deepfake images or videos are used to spoof a biometric system, such standardised metrics will become relevant.

\subsection{Discussion: Performance Metrics \& Standards}

This section provided a comprehensive summary of performance metrics used for evaluating and benchmarking binary classifiers. It is rare that all such metrics are used for a specific application. Instead, one or several are chosen based on specific needs. For a deepfake detection system as a binary classifier, many researchers have chosen to use overall metrics such as accuracy, AUC, EER and log loss, but the combination of precision, recall and F1-score is also common. Some deepfake-related challenges and competitions have introduced their own specific metrics, some of which will be described in Section~\ref{Sec:challenges}. The use of different performance metrics can make comparison of different reported results more difficult, so we hope the expected new ISO/IEC standard particularly ISO/IEC 4213 will help.

It is worth mentioning that, in addition to evaluating performance of deepfake detectors, the introduced performance metrics for evaluating binary classifiers can also be used to evaluate performance of deepfake generation methods by considering how deepfake detectors fail. For instance, organisers of the Voice Conversion Challenge 2018 and 2020 used this approach to benchmark how well voice conversion (VC) systems can generate high-quality fake speech samples.

Another point we would like to mention is that for deepfake videos there are two levels of performance metrics: those at the frame level (metrics of each frame), and those at the video level (metrics for the whole video). Generally speaking, the latter can be obtained by averaging the former for all frames, potentially following an adaptive weighting scheme, so that more important (key) frames will be counted more.
%%%%%%

\section{Deepfake-Related Datasets\label{sec:datasets}}

In this section, we cover all deepfake-related datasets we identified from the meta-review of deepfake-related survey papers, deepfake-related challenges, competitions and benchmarks covered, one online collections of deepfake-related datasets on GitHub\footnote{\url{https://github.com/592McAvoy/fake-face-detection\#user-content-i-dataset}}, and the co-authors' personal collections. Table~\ref{tab:datasets} shows basic information about these datasets. We explain them in four categories: deepfake image datasets, deepfake video datasets, deepfake audio/speech datasets, and hybrid deepfake datasets (mainly mixed image and video datasets).

Note that many datasets of real (authentic) media were also used by deepfake researchers for two purposes. First, any detectors would need both fake and real media to demonstrate their performance. Second, real media have also been used to train deepfake generators as the training set. In this section, we include only datasets containing deepfake media, some of which contain both deepfake and real media.

Some datasets, especially those created for deepfake-related challenges and competitions, have separate subsets for training and evaluation (testing) purposes. The split is necessary for such challenges and competitions, but not very useful for people who just want to use such datasets. Therefore, in this section when introducing such datasets we will ignore that level of details and focus on the total number of data including the number of real and fake samples.

\newcommand\FFIWtenK{$\text{\itshape FFIW}_{10K}$}
\begin{xltabular}{\linewidth}{>{\centering\arraybackslash}X>{\centering\arraybackslash}p{0.3\linewidth}c}
\caption{Deepfake-related datasets\label{tab:datasets}}\\\midrule
\textbf{Dataset} & \textbf{Size} & \textbf{Year} \\\midrule
\endfirsthead
\multicolumn{3}{c}{Table~\ref{tab:datasets}: Deepfake-related datasets (continued)}\\
\midrule
\textbf{Dataset} & \textbf{Size} & \textbf{Year}\\
\midrule
\endhead

SwapMe and FaceSwap dataset & 4310 images & 2017\\

Fake Faces in the Wild (FFW) dataset & 53,000 images (from 150 videos) & 2018\\

generated.photos datasets & 2.7 million images & Since 2018\\

MesoNet Deepfake Dataset & 19,509 images & 2018\\

100K-Generated-Images & 100,000 images & 2019\\

Ding et al.'s swapped face dataset & 420,053 images & 2019\\

iFakeFaceDB & 87,000 images & 2019\\

Faces-HQ & 40,000 images & 2019-20\\

CelebA-Spoof & 625,537 images & 2020\\

Diverse Fake Face Dataset (DFFD) & 299,039 images & 2020\\

\midrule

DeepfakeTIMIT & 620 videos & 2018\\

FaceForensics (FF) & 1,004 videos & 2018\\

UADFV dataset & 98 videos & 2018\\

DFDC (Deepfake Detection Challenge) preview dataset & 5,244 videos & 2019\\

FaceForensics++ (FF++) & 5,000 videos & 2019\\

Deep Fakes Dataset & 142 videos & 2019-20\\

Celeb-DF v1 & 1,203 videos & 2020\\ 

Celeb-DF v2 & 6,229 videos & 2020\\

DeepFake Detection (DFD) dataset & 3,363 videos & 2019\\

DeeperForensics-1.0 & 60,000 videos & 2020\\

DFDC (Deepfake Detection Challenge) full dataset & 128,154 videos & 2020\\

\FFIWtenK\ (Face Forensics in the Wild) dataset & 10,000 videos &2021\\

Korean DeepFake Detection Dataset (KoDF) & 37,942 videos &2021\\

VideoForensicsHQ & 1,737 videos & 2021\\

WildDeepfake & 7,314 face sequences (from 707 videos) & 2021\\

\midrule

Voice Conversion Challenge 2016 dataset & 2,160 ``real'' utterances + 918 ``fake'' utterances & 2016\\
Voice Conversion Challenge 2018 dataset & 1,392 ``real'' utterances + 1,190 ``fake'' utterances & 2018\\

ASVspoof 2019 dataset (Logical Access task) & 121,461 utterances & 2019\\

Voice Conversion Challenge 2020 dataset & 2,030 ``real'' utterances + 1,475 ``fake'' utterances & 2020\\

Baidu Research dataset & 134 utterances & 2020\\

ASVspoof 2021 Challenge -- Logical Access Database & 7.8 GB (compressed) &2021\\
ASVspoof 2021 Challenge -- Speech Deepfake Database & 34.5 GB (compressed) & 2021\\

\midrule

gpt-2-output-dataset & $250K\times 9$ documents & 2019\\

Grover dataset & 25,000 articles & 2019\\

TweepFake & 25,572 tweets from 23 bots and 17 human accounts & 2021\\

\midrule

NIST Open Media Forensics Challenge Datasets & Over 1,000 images and over 100 videos & 2020\\

ForgeryNet dataset & 2,896,062 images and 221,247 videos & 2021\\

\bottomrule
\end{xltabular}

\subsection{Deepfake Image Datasets}

\textbf{SwapMe and FaceSwap dataset}~\cite{zhou2017}: This dataset contains 4,310 images, including 2,300 real images and 2,010 fake images created using FaceSwap\footnote{\url{https://github.com/MarekKowalski/FaceSwap/}} and the SwapMe iOS app (now discontinued).

\textbf{Fake Faces in the Wild (FFW) dataset}~\cite{khodabakhsh2018}: This dataset contains 131,500 face images, including 78,500 images extracted from 150 videos in the FaceForensics dataset and 53,000 images extracted from 150 fake videos collected from YouTube.

\textbf{generated.photos datasets}\footnote{\url{https://generated.photos/datasets}}: This is a number of commercial datasets provided by the Generated Media, Inc., with up to nearly 2.7 million synthetic face images generated by StyleGAN. A free edition with 10,000 128x128 synthetic images is made available for academic research. The website also provides an interactive face generator\footnote{\url{https://generated.photos/face-generator/new}} and an API\footnote{\url{https://generated.photos/api}}. The generated.photos datasets have a good diversity: five age groups (infants, children, youth, adults, middle-aged), two genders (male and female), four ethnicities (white, black, Latino, Asian), four eye colours (brown, grey, blue, green), four hair colours (brown, black, blond, gray), three hair length (short, medium, long), facial expressions, three head poses (front facing, left facing, right facing), two emotions (joy and neutral), two face styles (natural, beautified). (According to a number of research papers we read, an earlier 100K-Faces dataset was released by generated.photos for academic research in 2018, which was used by many researchers. This dataset is not currently available any longer.)

\textbf{MesoNet Deepfake Dataset}~\cite{afchar2018}: This dataset includes 19,457 face images, including 7,948 deepfake images generated from on 175 forged videos collected online and 11,509 real face images collected from various online sources. (Table~2 of the paper shows the dataset size is 19,509, but the dataset downloaded from pCloud contains just 19,457 images.)

\textbf{100K-Generated-Images}~\cite{karras2019style}: This dataset includes 100,000 synthesised face, bedroom, car and cat images by a GAN generator trained based on real images in the FFHQ\footnote{\url{https://github.com/NVlabs/ffhq-dataset}} and LSUN\footnote{\url{https://github.com/fyu/lsun}} datasets (three object types -- bedrooms, cars and cats -- for the latter). Note that the name ``100K-Generated-Images'' was not a proper one as the authors~\cite{karras2019style} just used this to name a sub-folder of their Google Drive shared space, but it was used in one of the survey papers~\cite{TWPW2020}.

\textbf{Ding et al.'s swapped face dataset}~\cite{ding2020swapped}: This dataset contains 420,053 images of celebrities, including 156,930 real ones downloaded using Google Image API and 263,123 fake face-swapped ones created using two different methods (Nirkin's method and Auto-Encoder-GAN)

\textbf{iFakeFaceDB}~\cite{neves2020}: This dataset includes 87,000 224x224 face images, generated by processing some StyleGAN-generated synthetic images using the GAN-fingerprint Removal approach (GANprintR) proposed by \citeauthor{neves2020}. It is the replaced version of the \textbf{FSRemovalDB} dataset, which contains 150,000 face images generated using an earlier version of GANprintR.

\textbf{Faces-HQ}~\cite{durall2019}: This dataset includes 40,000 images, half real and half deepfake. The images were collected from four sources: the CelebA-HQ dataset\footnote{\url{https://drive.google.com/open?id=0B4qLcYyJmiz0TXY1NG02bzZVRGs}}, the Flickr-Faces-HQ dataset\footnote{\url{https://github.com/NVlabs/ffhq-dataset}}, the 100K-Faces dataset\footnote{\url{https://generated.photos/}} (not available any longer, see the description of generated.photos datasets), and \href{https://thispersondoesnotexist.com/}{thispersondoesnotexist.com}.

\textbf{CelebA-Spoof}~\cite{zhang2020}: This dataset includes 625,537 synthesised face images of 10,177 celebrities, with 43 rich attributes on face, illumination, environment and spoof types. The real images were selected from the CelebA dataset\footnote{\url{http://mmlab.ie.cuhk.edu.hk/projects/CelebA.html}}. The 43 attributes include 40 for real images, covering all facial components and accessories (e.g., skin, nose, eyes, eyebrows, lip, hair, hat, eyeglass), and 3 for fake images, covering spoof types, environments and illumination conditions.

\textbf{Diverse Fake Face Dataset (DFFD)}~\cite{Dang_2020_CVPR}: This dataset contains 299,039 images, including 58,703 real images sampled from three datasets (FFHQ\footnote{\url{https://github.com/NVlabs/ffhq-dataset}}, CelebA\footnote{\url{http://mmlab.ie.cuhk.edu.hk/projects/CelebA.html}} and FaceForensics++\footnote{\url{https://github.com/ondyari/FaceForensics}}) and 240,336 fake ones in four main facial manipulation types (identity swap, expression swap, attribute manipulation, and entire synthesis). The images cover two genders (male and female), a wide age groups (the majority between 21 and 50 years old), and both low- and high-quality levels.

\subsection{Deepfake Video Datasets}

\textbf{DeepfakeTIMIT}~\cite{korshunov2019}: This dataset contains 620 deepfake face videos, generated by face swapping without manipulation of audio, covering 32 subjects and two quality levels (high and low).

\textbf{FaceForensics} (FF)~\cite{rossler2018}: This dataset contains 1,004 face videos with over 500,000 frames, covering various quality levels and two types of facial manipulation. This dataset is now replaced by the larger FaceForensics++ dataset (see below).

\textbf{FaceForensics++} (FF++)~\cite{RCVRTN2019}: This dataset contains 5,000 face videos with over 1.8 million manipulated frames, including 1,000 real videos (with 509,914 frames) downloaded from YouTube, and 4,000 fake videos created using four face manipulation methods (Deepfakes, Face2Face, FaceSwap and NeuralTextures). The videos cover two genders (male and female), and three quality levels (VGA/480p, HD/720p,  and FHD/1080p).

\textbf{UADFV dataset}~\cite{li2018}: This dataset contains 98 face videos, half (49) are real ones downloaded from Youtube, and the other half are fake ones generated using the FakeApp mobile application (which is now discontinued). The video dataset was created to used to demonstrate a deepfake video detection method based on detection of eye blinking behaviours, so all videos contain at least one eye-blinking event. All fake videos were created by swapping the original face in each of the real videos with the face of the actor Nicolas Cage\footnote{\url{https://en.wikipedia.org/wiki/Nicolas_Cage}}, thus, only one subject is represented.

\textbf{Deep Fakes Dataset}~\cite{ciftci2020}: This dataset contains 142 ``in the wild'' deepfake portrait videos, collected from a range of online sources including news articles, online forums, mobile apps, and research presentations. The videos are diverse, covering the source generative model, resolution, compression, illumination, aspect-ratio, frame rate, motion, pose, cosmetics, occlusion, content, and context.

\textbf{DFDC (Deepfake Detection Challenge) preview dataset}~\cite{dolhansky2020}: This dataset contains 5,244 face videos of 66 subjects with both face and voice manipulation. It was released as a preview of the full dataset of the 2020 Deepfake Detection Challenge (DFDC, see below).

\textbf{Celeb-DF v1}\footnote{\url{https://github.com/yuezunli/celeb-deepfakeforensics/tree/master/Celeb-DF-v1}}: This dataset contains 1,203 face videos of celebrities, including 408 real videos collected from YouTube with subjects of different ages, ethic groups and genders, and 795 deepfake videos synthesised from these real videos.

\textbf{Celeb-DF v2}~\cite{LYSQL2020}: This dataset contains 6,229 face videos of celebrities, including 590 real videos collected from YouTube with subjects of different ages, ethic groups and genders, and 5,639 deepfake videos synthesised from these real videos.

\textbf{DeepFake Detection (DFD) Dataset}~\cite{googleai2019}: This dataset contains 3,363 face videos, covering 28 subjects, gender, and skin colour. It was created as a joint effort between two units of Google, Inc.: Google AI\footnote{\url{https://ai.googleblog.com/}} and JigSaw\footnote{\url{https://jigsaw.google.com/}}.

\textbf{DeeperForensics-1.0}~\cite{JLWQL2020}: This dataset contains 60,000 indoor face videos (with 17.6 million frames) generated by face swapping, covering 100 subjects, four skin tones (white, black, yellow, brown), two genders (male and female), different age groups (20-45), 26 nationalities, 7 different angles, 8 face expressions, and different head poses.

\textbf{DFDC (Deepfake Detection Challenge) full dataset}~\cite{dolhansky2020}: This dataset contains 128,154 face videos of 960 subjects, including 23,654 real videos  from 3,426 paid actors and 104,500 deepfake videos created using eight different methods (DF-128, DF-256, MM/NN face swap, NTH, FSGAN, StyleGAN, refinement, and audio swap).

\textbf{\FFIWtenK\ (Face Forensics in the Wild) dataset}~\cite{zhou2021}: This dataset contains 10,000 high-quality forgery videos, with video- and face-level annotations. The dataset focuses on a more challenging case for forgery detection: each video involves one to 15 individuals, but only some (a minority of) faces are manipulated.

\textbf{Korean DeepFake Detection Dataset (KoDF)}~\cite{kwon2021}: This dataset contains 37,942 videos of paid subjects (395 Koreans and 8 Southeastern Asians), including 62,166 real videos and 175,776 fake ones created using six methods -- FaceSwap, DeepFaceLab, FSGAN, First Order Motion Model (FOMM), Audio-driven Talking Face HeadPose (ATFHP) and Wav2Lip. The videos cover a balanced gender ratio and a wide range of age groups.

\textbf{VideoForensicsHQ}~\cite{fox2021}: This dataset contains 1,737 videos with 1,666,816 frames, including 1,339,843 real frames and 326,973 fake frames generated using the Deep Video Portraits (DVP)~\cite{kim2018} method. The original videos were obtained from three sources: the dataset used in \cite{kim2019neural}, the Ryerson Audio-Visual Database of Emotional Speech and Song (RAVDESS)~\cite{livingstone2018ryerson}, and YouTube. Most videos have a resolution of 1280×720.

\textbf{WildDeepfake}~\cite{zi2020}: This dataset contains 7,314 face sequences extracted from 707 deepfake videos that were collected completely from the Internet. It covers diverse scenes, multiple persons in each scene and rich facial expressions. Different from other deepfake video datasets, WildDeepfake contains only face sequences not the full videos. This makes the dataset more like between an image dataset and a video one. We decided to keep it in the video category since the selection process was still more video-focused.

\subsection{Deepfake Audio/Speech Datasets}

Voice conversion (VC) is a technology that can be used to modify an audio and speech sample so that it appears as if spoken by a different (target) person than the original (source) speaker. Obviously, it can be used to generate deepfake audio/speech samples. The biennial Voice Conversion Challenge\footnote{\url{http://www.vc-challenge.org/}} that started in 2016 is a major challenge series on VC. Datasets released from this challenge series are very different from other deepfake datasets: the deepfake data is not included in the original dataset created by the organisers of each challenge, but in the participant submissions (which are retargeted/fake utterances produced by VC systems built by participants). The challenge datasets also include the evaluation (listening-based) results of all submissions. Some fake utterances may be produced by DL-based VC systems, so we consider all datasets from this challenge series relevant for our purpose of this survey.

\textbf{Voice Conversion Challenge 2016 database}~\cite{toda16_interspeech}: The original dataset created by the challenge organisers was derived from the DAPS (Device and Produced Speech) Dataset~\cite{mysore2015}. It contains 216 utterances (162 for training and 54 for testing) per speaker from 10 speakers. Participating teams (17) developed their own VC systems for all 25 source-target speaker pairs, and then submitted generated utterances for evaluation. At least six participating teams used DL-related techniques (LSTM, DNN) in their VC systems (see Table~2 of the result analysis paper\footnote{\url{http://www.vc-challenge.org/vcc2016/papers/SSW9_VCC2016_Results.pdf}}), so the submitted utterances can certainly be considered deepfakes.

\textbf{Voice Conversion Challenge 2018 database}~\cite{Lorenzo-Trueba2018}: The original dataset created by the challenge organisers was also based on the DAPS dataset. It contains 116 utterances (81 for training and 35 for testing) per speaker from 12 speakers in two different tasks (called Hub and Spoke). Participating teams (23 in total, all for Hub and 11 for Spoke) developed their own VC systems for all 16 source-target speaker pairs, and then submitted generated utterances for evaluation. Comparing with the 2016 challenge, more participating teams used DL-related techniques (e.g., WaveNet, LSTM, DNN, CycleGAN, DRM -- deep relational models, and ARBM -- adaptive restricted Boltzmann machines) in their VC systems.

\textbf{Voice Conversion Challenge 2020 database}~\cite{YHTYDKLT2020}: This dataset is based on the Effective Multilingual Interaction in Mobile Environments (EMIME) dataset\footnote{\url{https://www.emime.org/participate/emime-bilingual-database.html}}, a bilingual (Finnish/English, German/English, and Mandarin/English) database. It contains 145 utterances (120 for training and 25 for testing) per speaker from 14 speakers for two different tasks (with $4\times 4$ and $4\times 6$ source-target speaker pairs, respectively). Participating teams (33 in total, out of which 31 for Task 1 and 28 for Task 2) developed their own VC systems for all source-target speaker pairs, and then submitted generated utterances for evaluation. Comparing with the 2018 challenge, DL-based VC systems were overwhelmingly used by almost all participating teams (WaveNet and WaveGAN among the most used DL-based building blocks).

A major set of deepfake speech datasets were created for the \textbf{ASVspoof} (Automatic Speaker Verification Spoofing and Countermeasures) Challenge\footnote{\url{https://www.asvspoof.org/}} (2015-2021, held biannually). The datasets for the 2019 and 2021 contain speech data that can be considered deepfakes.

\textbf{ASVspoof 2019 Challenge database}~\cite{WANG2020101114}: This dataset is based on the Voice Cloning Toolkit (VCTK) corpus\footnote{\url{https://doi.org/10.7488/ds/1994}}, a multi-speaker English speech database captured from 107 speakers (46 males and 61 females). Two attack scenarios were considered: logical access (LA) involving spoofed (synthetic or converted) speech, and physical access (PA) involving replay attacks of previously recorded bona fide recordings). For our purpose in this survey, the LA scenario is more relevant. The LA part of the dataset includes 12,483 bona fide (real) utterances and 108,978 spoofed utterances. Some of the spoofed speech data for the LA scenario were produced using a generative model involving DL-based techniques such as long short-term memory (LSTM)\footnote{\url{https://www.cs.toronto.edu/~graves/phd.pdf}}, WaveNet~\cite{oord2016}, WaveRNN~\cite{kalchbrenner2018}, WaveCycleGAN2~\cite{tanaka2019}. Note that the challenge organisers did not use the term ``deepfake'' explicitly, despite the fact that the DL-generated spoofed speech data can be considered as deepfakes.

\textbf{ASVspoof 2021 Challenge -- Logical Access Database}~\cite{delgado_hector_2021_1}: This dataset contains bona fide and spoofed speech data for the logical access (LA) task. The challenge is still ongoing and we did not find a detailed paper on the dataset, so cannot include more details other than its size (7.8 GB after compression). Although we did not see details of the generative algorithms used to produce spoofed speech data, we believe similar DL-based algorithms were used like for the 2019 challenge.

\textbf{ASVspoof 2021 Challenge -- Speech Deepfake Database}~\cite{delgado_hector_2021_2}: In 2021, the challenge included an explicitly defined track on deepfake, but the task description suggests that the organisers of the challenge considered a broader definition of the term ``deepfake'' by looking at spoofing human listeners rather than ASV (Automatic Speaker Verification) systems. The size of the dataset is 34.5 GB after compression.

Possibly because of the long history and wide participation of the community in the ASVspoof challenges for creating the dedicated datasets, there are very few other deepfake audio/speech datasets. One such dataset was created by a group of researcher from Baidu Research~\cite{arik2018}. This dataset was created to demonstrate a proposed voice cloning method. It is relatively small, and contains 134 utterances, including 10 real ones, 120 cloned ones, and 4 manipulated ones. Another dataset was created by Google AI and Google News Initiative\footnote{\url{https://www.blog.google/outreach-initiatives/google-news-initiative/advancing-research-fake-audio-detection/}}, but it was made part of the ASVspoof 2019 dataset. This dataset contains thousands of phrases spoken by 68 synthetic ``voices'' covering a variety of regional accents.

\subsection{Hybrid Deepfake Datasets}

\textbf{NIST OpenMFC (Open Media Forensics Challenge) Datasets}\footnote{\url{https://mfc.nist.gov/\#pills-data}}: These datasets were created by the DARPA Media Forensics (MediFor) Program\footnote{\url{https://www.darpa.mil/program/media-forensics}} for the 2020 OpenMFC\footnote{\url{https://mfc.nist.gov/}}. There are two GAN-generated deepfake datasets, one with more than 1,000 deepfake images and the other with over 100 deepfake videos. The datasets were made available to registered participants of the competition only.

\textbf{ForgeryNet}~\cite{He_2021_CVPR}: This dataset is named as ``a versatile benchmark for comprehensive forgery analysis''. It contains 2,896,062 images and 221,247 videos, including 1,457,861 fake images and 121,617 fake videos. The videos and images cover seven image-level and eight video-level manipulation approaches, 36 different types of perturbations and more mixed perturbations, and a large number of annotation labels (6.3 million classification labels, 2.9 million manipulated area annotations and 221,247 temporal forgery segment labels). The dataset is being used for supporting the Face Forgery Analysis Challenge 2021\footnote{\url{https://competitions.codalab.org/competitions/33386}} at the SenseHuman 2021 (3rd Workshop on Sensing, Understanding and Synthesizing Humans)\footnote{\url{https://sense-human.github.io/}}, co-located at the ICCV 2021 conference\footnote{\url{http://iccv2021.thecvf.com/}}.

\subsection{A Deepfake Dataset Generator}

\textbf{DatasetGAN}~\cite{zhang2021}: This is not actually a dataset per se, but a system for producing large datasets more automatically, including generating deepfake datasets. One may argue the automatically generated datasets are fake since they are not produced from real-world scenes.

\subsection{Subjective Quality of Deepfakes in Different Databases\label{Subsec:datasets_subjective_quality}}

As mentioned in Section~\ref{sec:metrics_summary}, subjective quality evaluation is necessary to evaluate the realness, realisticness, and naturalness of deepfake media. While there has been very limited work on this topic, in 2020, \citet{JLWQL2020} conducted a user study on realness of deepfake videos. They recruited 100 professional participants (most of whom are computer vision researchers), who were asked to evaluate the realness of 30 randomly selected videos from 7 deepfake video datasets (DeeperForensics-1.0, UADFV, DeepFake-TIMIT, Celeb-DF, FaceForensics++, Deep Fake Detection, and DFDC). Participants were asked to respond to the statement ``The video clip looks real." and gave scores following a five-point Likert scale (1 -- clearly disagree, 2 -- weakly disagree, 3 -- borderline, 4 -- weakly agree, 5 -- clearly agree). Table~\ref{tab:datasets_MOS} shows the results. Interestingly, we can see a huge difference between the realness levels of different datasets. What is probably quite surprising is that FaceForensics++, one of the most widely used deepfake datasets, has a very low MOS score and less than 9\% of participants considered the 30 selected videos as real.

\begin{table}[!h]
\begin{center}
\caption{Human-judged subjective quality (realness) of deepfake videos in 7 datasets. The MOS scores were not reported by \citeauthor{JLWQL2020}, but calculated by us based on the raw data shown in Table 3 of \cite{JLWQL2020}.}
\label{tab:datasets_MOS}
\begin{tabular}{ccc}
\toprule
Dataset & MOS & 4+ ratings (\%)\\
\midrule
DeeperForensics-1.0 & 3.806 & 64.1\%\\
Celeb-DF & 3.723 & 61.0\%\\
DFDC & 2.539 & 23\%\\
Deep Fake Detection & 2.518 & 21.9\%\\
UADFV & 2.249 & 14.1\%\\
DeepFake-TIMIT & 2.205 & 12.3\%\\
FaceForensics++ & 1.874 & 8.4\%\\
\bottomrule
\end{tabular}
\end{center}
\end{table}

\subsection{Discussion: Datasets}
\label{sec:metrics_summary}

Among all deepfake image and video datasets, a significant majority are about face images and videos. This is not surprising since face swapping, face attribution manipulation, and fully synthesised face images are among the hottest topics within deepfake research and real-world applications. We hope more non-face deepfake image and video datasets can be produced to support a broader range of research activities on deepfake.

The subjective quality results shown in Table~\ref{tab:datasets_MOS} indicate that it is important to check realness of deepfake media to support any performance evaluation or comparison. To ensure that the quality evaluation of datasets is fair, transparent and reliable, standard procedures need defining and a common pool of qualified human experts should be used.

Many authors of deepfake-related datasets attempted to classify such datasets into different generations. Chronologically speaking, we could broadly split such datasets into two generations: before 2019 and since 2019. Typically, datasets created before 2019 are relatively less advanced and smaller, while those created after 2019 tend to be larger, more diverse (i.e., covering more attributes), and of higher quality (i.e., produced by more advanced generative models). This can also be seen from the data in Table~\ref{tab:datasets_MOS}, in which the top two datasets (DeeperForensics-1 and Celeb-DF) fall within the new  generation (2020), while others belong to the old generation. In addition to the two generations, a newer generation has also emerged in 2021: a number of very recent datasets started focusing on more realistic deepfakes (i.e., in the wild) or more specified areas of deepfakes (e.g., $\text{\itshape FFIW}_{10K}$ focusing on multiple faces in the same video, and KoDF focusing on Korean faces). This trend shows that the deepfake research community has grown significantly in the past few years so that narrower topics have also started gaining attention and interest from some researchers.
%%%%
\section{Deepfake-Related Challenges, Competitions \& Benchmarks\label{Sec:challenges}}

This section reviews initiatives aiming to advance the state-of-the-art of detection and generation of synthetic or manipulated media (such as video, image and audio) via competitions or challenges open to the public, and on-going benchmarks tackling specific problems. 

\subsection{Detection of Manipulated Media}

\begin{figure*}[!htb]
\centering
\includegraphics[width=0.9\linewidth]{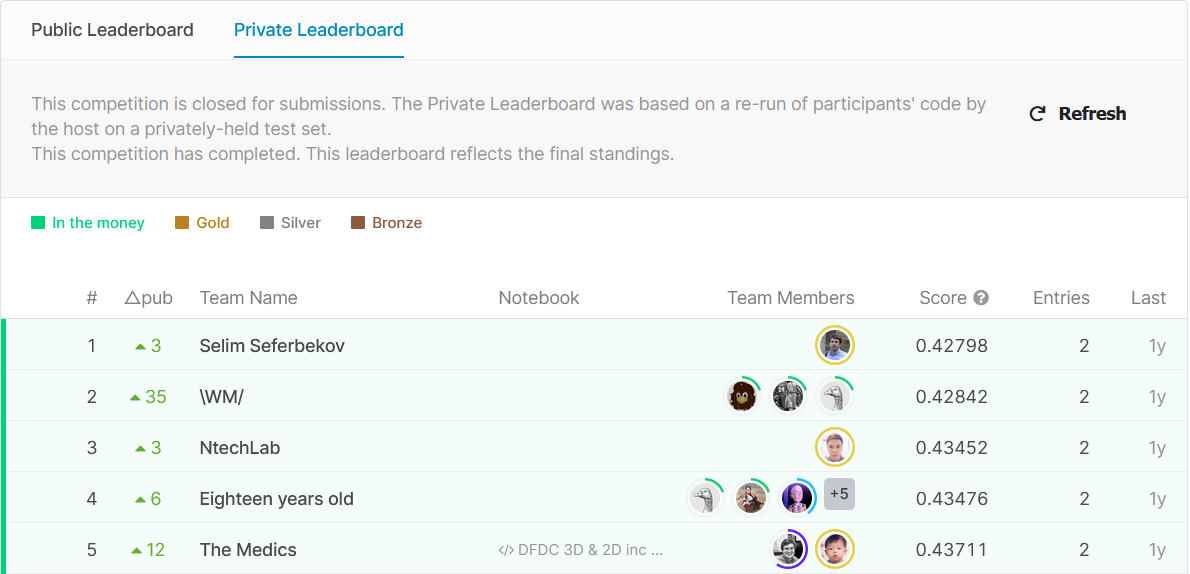}
\caption{Screenshot of leaderboard with top five finalists of the DFDC competition.}
\label{fig:DFDC}
\end{figure*}

The Deepfake Detection Challenge (DFDC)\footnote{\url{https://www.kaggle.com/c/deepfake-detection-challenge}} was an initiative promoted by an AI and Media Steering Committee\footnote{\url{https://www.partnershiponai.org/ai-and-media-integrity-steering-committee/}}, including BBC, Facebook, Amazon, Microsoft and New York Times, and some universities around the world including the University of Oxford. The competition remained open from 5 September 2019 till 31 March 2020, and involved 3 stages. At first, the DFDC preview dataset was released. At a later stage, the DFDC full dataset was also made available to the 2,114 participants of the competition incorporating face and audio swap techniques for generation of deepfake content. 
At the final stage, the submitted models were evaluated using a test dataset (referred to as the ``black box dataset'') of 10,000 videos which included \emph{in-the-wild} deepfake videos.
The best performance on the black box dataset had an accuracy of 65.18\%, according to the released results~\cite{metaai2020}. Submissions were ranked\footnote{\url{https://www.kaggle.com/c/deepfake-detection-challenge/leaderboard}} according to the overall log loss score, as defined in Eq.~\eqref{eq:LL}. All top five ranked models (the winner had the lowest overall log loss) are available on GitHub. Results indicate how challenging the detection of deepfake is since the best accuracy was low and ``\textit{many  submissions were simply random}'', according to \citet{DBPLHWF2020}. Figure~\ref{fig:DFDC} shows a screenshot of the leaderboard with the five finalists. The first top ranked model used MTCNN (Multi-tasked Cascaded Convolutional Network), the second used WS-DAN (Weakly Supervised Data Augumentation Network), and the third used the EfficientNetB7 architecture.
Meta compiling the common themes observed in the winning models, they were: clever augmentations, architectures, and absence of forensics methods. Moving forward, they called for ``\textit{solutions that go beyond analysing images and video. Considering context, provenance, and other signals may be the way to improve deepfake detection models}''.

The Automatic Speaker Verification Spoofing And Countermeasures Challenge Workshop (ASVspoof)\footnote{\url{https://www.asvspoof.org/}} has been running biennially since 2015. This competition is organised by an international consortium that includes Inria and EURECOM (France), University of Eastern Finland, National Institute of Informatics (Japan), and Institute for Infocomm Research (Singapore). This year the ASVspoof challenge includes, for the first time, a sub-challenge focused on \emph{Speech DeepFake} where the envisioned use case is an adversary trying to fool a human listener. The metric used for evaluating performance of submitted solutions (i.e., classifiers) is EER. Four baseline solutions\footnote{\url{https://competitions.codalab.org/competitions/32345\#learn_the_details}} (also called ``countermeasures''), each using a different technique, were made available to participants with their corresponding EER metric values. The ASVspoof 2021 Speech Deepfake Database containing audio recordings with original and spoofed utterances has also been made available. The competition involves three phases\footnote{\url{https://competitions.codalab.org/competitions/32345\#phases}}: a progress phase, an evaluation phase and a post-evaluation phase; it is unclear how teams move from one phase to the next. 
More information about the 2021 competition is available in the published evaluation plan~\cite{DEKLLNPSTWY2021}. The organisers of the competition noted that they opted for the EER as the performance evaluation metric for countermeasures submitted to the speech deepfake task for legacy reasons. They acknowledged, however, that ``\textit{EER reporting is deprecated}'' by the ISO/IEC 19795-1:2021\footnote{\url{https://www.iso.org/standard/73515.html}} standard. Despite the fact that only the 2021 ASVspoof competition contained a track explicitly related to deepfake, some data in the ASVspoof 2019 dataset (Logical Access task) used for the 2019 competition was generated using DL-based algorithms as mentioned in Section~\ref{sec:datasets}. We expect that this also holds for the ASVspoof 2021 dataset (Logical Access task). The ASVspoof 2019 competition used the EER as secondary metric; the primary performance metric used was the tandem detection cost function (t-DCF)~\cite{TWVSDNYEKL2019}. According to its evaluation plan~\cite{YTSDWEKLVN2019}, t-DCF assesses the performance of the whole tandem system whereby ``\textit{a CM [countermeasure] serves as a `gate' to determine whether a given speech input originates from a bona fide (genuine) user, before passing it the main biometric verifier (the ASV system)}''. It is calculated according to Eq.~\eqref{eq:tDCF}, where $P^{\text{cm}}_{\text{miss}}(s)$ and $P^{\text{cm}}_{\text{fa}}(s)$ are, respectively, ``\textit{the miss rate and the false alarm rate of the CM system at threshold s}''.

\begin{align}
    \text{t-DCF} & = C_{1} P^{\text{cm}}_{\text{miss}}(s) + C_{2}P^{\text{cm}}_{\text{fa}}(s) \label{eq:tDCF} \\
    P^\text{{cm}}_\text{{miss}}(s) & = \frac{\#\{\text{bona\ fide\ trials\ with\ CM\ score} \leq s\}}{\#\{\text{Total\ bona\ fide\ trials\}}}\nonumber \\
    P^\text{{cm}}_\text{{miss}}(s) & = \frac{\#\{\text{spoof\ trials\ with\ CM\ score} > s\}}{\#\{\text{Total\ spoof\ trials\}}} \nonumber
\end{align}

For further information about Eq.~\eqref{eq:tDCF}, including constants $C_{1}$ and $C_{2}$, please refer to the ASVspoof 2019 evaluation plan~\cite{YTSDWEKLVN2019}.

An implementation of the t-DCF metric has been made available by the ASVspoof 2019's organisers in Python\footnote{\url{https://www.asvspoof.org/resources/tDCF_python_v2.zip}} and Matlab\footnote{\url{https://www.asvspoof.org/resources/tDCF_matlab_v2.zip}} formats.

\begin{figure*}[!htb]
\centering
\includegraphics[width=\linewidth]{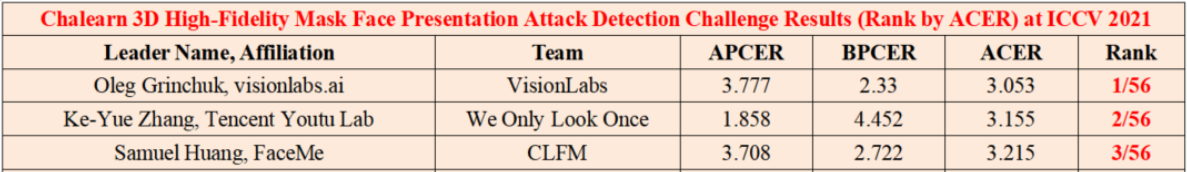}
\caption{Screenshot of leaderboard with top three finalists of the Face Anti-spoofing Challenge 2021 competition.}
\label{fig:FAS}
\end{figure*}

The Face Anti-spoofing (Presentation Attack Detection) Challenge\footnote{\url{https://sites.google.com/qq.com/face-anti-spoofing/welcome/challengeiccv2021}} started in 2019. Its first two editions were held at the 2019 and 2020 IEEE/CVF Conference on Computer Vision and Pattern Recognition (CVPR 2020), respectively. Its third edition was moved to be co-located with the 2021 IEEE/CVF International Conference on Computer Vision (ICCV 2021). This competition series was organised by a group of researchers from academia and industry in China, Mexico, Spain, Finland and the US.
The 2021 competition was focused on 3D high-fidelity mask attacks, and followed a 2-phased\footnote{\url{https://competitions.codalab.org/competitions/30910\#phases}} process. The first phase is the ``development phase''; it started in April 2021 when the CASIA-SURF HiFiMask dataset\footnote{\url{https://sites.google.com/qq.com/face-anti-spoofing/dataset-download/casia-surf-hifimaskiccv2021}} 
was released to participants. The second phase is the ``final ranking phase'' (June 2021), when the competition ended. The competition adopted the following  performance metrics for evaluation\footnote{\url{https://sites.google.com/qq.com/face-anti-spoofing/evaluation}} of the solutions submitted: attack presentation classification error rate (APCER), normal/bona fide presentation classification error rate (NPCER/BPCER), and average classification error rate (ACER), in accordance with the ISO/IEC 30107-3:2017\footnote{\url{https://www.iso.org/standard/67381.html}} standard. Figure~\ref{fig:FAS} provides the leaderboard for the top three solutions.

\begin{figure*}[!htb]
\centering
 \begin{subfigure}[b]{0.22\textwidth}
     \centering
     \includegraphics[width=\textwidth]{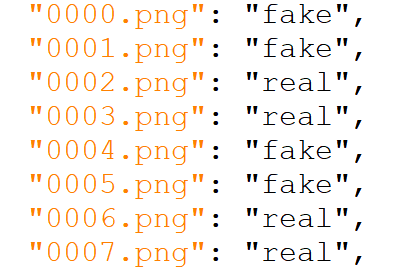}
     \caption{Example submission.}
     \label{fig:SUB}
 \end{subfigure}
 \hfill
 \begin{subfigure}[b]{0.77\textwidth}
     \centering
     \includegraphics[width=\textwidth]{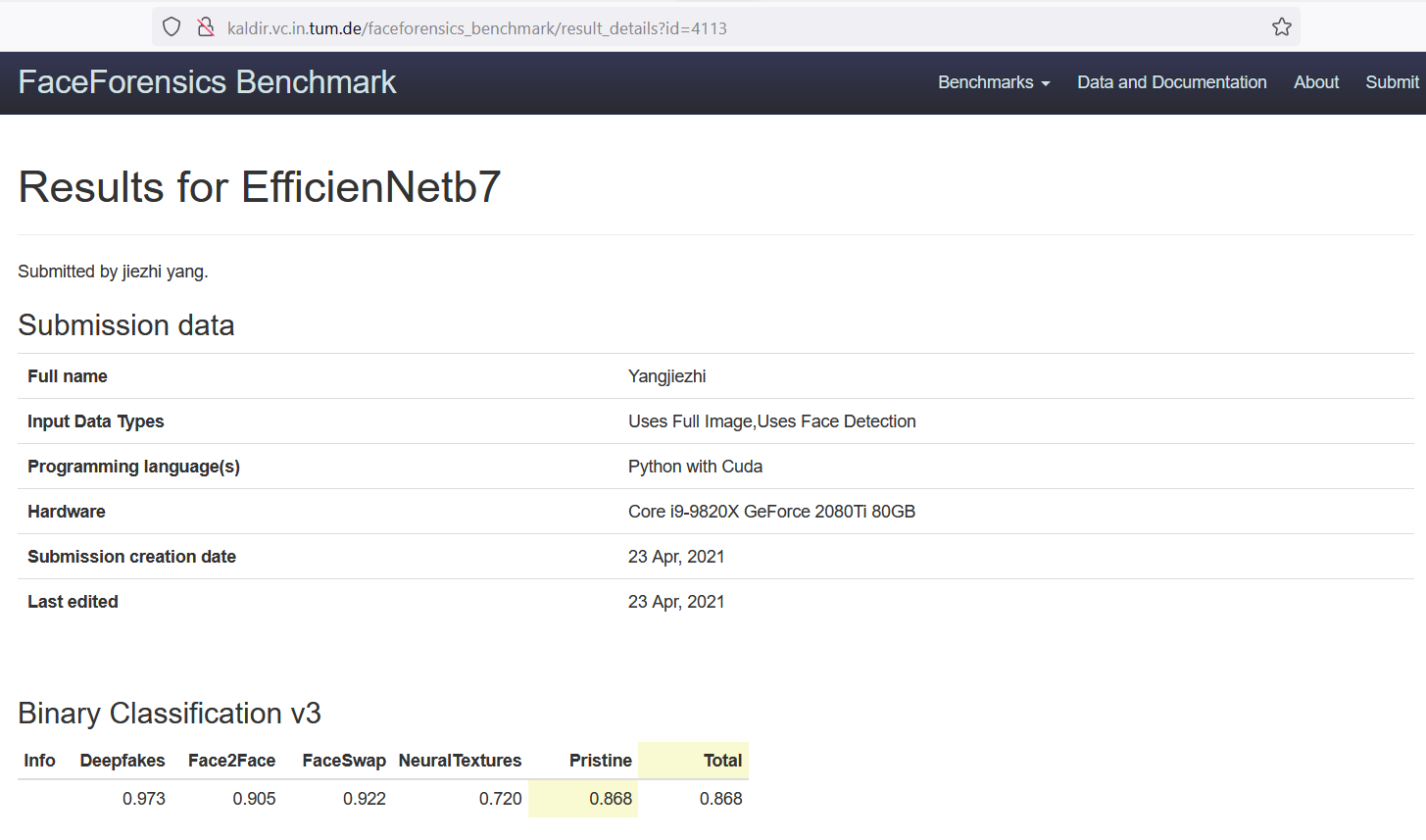}
     \caption{Example of submission result.}
     \label{fig:RES}
 \end{subfigure}
\caption{Illustration of the FaceForensics Benchmark in terms of submission and result.}
\label{fig:FFB}
\end{figure*}

The FaceForensics Benchmark\footnote{\url{http://kaldir.vc.in.tum.de/faceforensics_benchmark/}} is an on-going automated benchmark for detection of face manipulation. The organisers of the benchmark made the FaceForensics++ dataset 
available for training.
Manipulated videos (4,000 in total) were created using four techniques, i.e., two computer graphics-based approaches (Face2Face and FaceSwap) and two learning-based approaches (DeepFakes and Neural Textures). The deepfakes videos were generated using a slightly modified version of FaceSwap\footnote{\url{https://faceswap.dev/}}, and the Neural Textures videos were created using the approach proposed by~\citet{TZN2019}. The benchmark test dataset is created from the collection of 1,000 images randomly selected from either the manipulation methods or the original videos~\cite{RCVRTN2019}. Participants have to submit results to the benchmark, rather then code like other competitions; this is illustrated in Figure~\ref{fig:SUB}. The outcome of a submission is illustrated in Figure~\ref{fig:RES}, where the scores are a measure of accuracy (Eq.~\eqref{eq:accuracy}). 

The Open Media Forensics Challenge (OpenMFC, formerly DARPA MFC)\footnote{\url{https://mfc.nist.gov/}} is an annual image and video forensics evaluation aiming to facilitate development of multimedia manipulation detection systems. It has been organised annually\footnote{\url{https://www.nist.gov/itl/iad/mig/open-media-forensics-challenge}} starting from 2017 under the name of DARPA MFC. In 2020, the National Institute of Standards and Technology (NIST) initiated the \emph{OpenMFC} as a new evaluation platform, based on their previous experiences with the DARPA MFC series, to make the participation more convenient for all researchers. In OpenMFC 2020, two deepfake-related tasks were included for the first time: Image GAN Manipulation Detection (IGMD) and Video GAN Manipulation Detection (VGMD). 
The organisers provided an image evaluation dataset for the IGMD task, containing 1,000 images from over 200 image journals\footnote{``\textit{This is an automatically generated manipulation history graph log of media file manipulations with automatic output manipulation masks from a detector algorithm. Each journal tracks the media manipulations and software according to NIST manipulation data collection guidelines.}''~\cite{NISTIR8377}.}, and a video evaluation dataset for the VGMD task, including over 100 test videos. Furthermore, they provided the datasets\footnote{\url{https://mfc.nist.gov/\#pills-data}} used in the previous MFC challenges as development datasets. The challenge is composed of two main phases for development and evaluation, respectively, and a pre-challenge phase for quality control testing. For evaluation of submissions, 
AUC-ROC is used as the primary metric. Furthermore, CDR@FAR, where CDR refers to correct detection rate or TPR (Eq.~\eqref{eq:TPR}) and FAR refers to false alarm rate or FPR (Eq.~\eqref{eq:FPR}), is also used as a metric~\cite{NIST2021}.

The DeeperForensics Challenge 2020\footnote{\url{https://competitions.codalab.org/competitions/25228}} is a deepfake face detection challenge held at the 2020 ECCV SenseHuman Workshop\footnote{\url{https://sense-human.github.io/}}. The challenge used the DeeperForensics1.0 dataset.

The organisers provided a hidden test dataset to better simulate real-world scenarios. The challenge involved two phases: the ``development phase'' that started in August 2020 allowing 100 successful submissions, and the ``final test phase'' that started in October 2020 allowing 2 successful submissions until the end of the month. The submissions were evaluated using the binary cross-entropy loss (BCELoss) metric, calculated according to Eq.~\eqref{eq:BCE}, where $N$ is the  number of videos in the hidden test set, $y_{i}$ is the ground truth label of video $i$ (fake:1, real:0), and $p(y_{i})$ is the predicted probability that video $i$ is fake. 

\begin{align}
\text{BCELoss} & = -\frac{1}{N}\sum_{i=1}^{N} [y_{i}\times \log(p(y_{i})) + A \label{eq:BCE} \\
A & = (1 - y_{i})\times \log(1 - p(y_{i}))] \nonumber
\end{align}

Results\footnote{\url{https://competitions.codalab.org/competitions/25228\#results}} of the competition were discussed by~\citet{JGWLLLYXXCZLCYDLHCJLT2021}. The top solution used three models, i.e., EfficientNet-B0, EfficientNet-B1 and EfficientNet-B2, for classification. The second top used EfficientNet-B5 for both an image-based model and a video-based model. The third ranked solution used a 3D convolutional neural network (3DCNN).

\begin{figure*}[!htb]
    \centering
    \includegraphics[width=\textwidth]{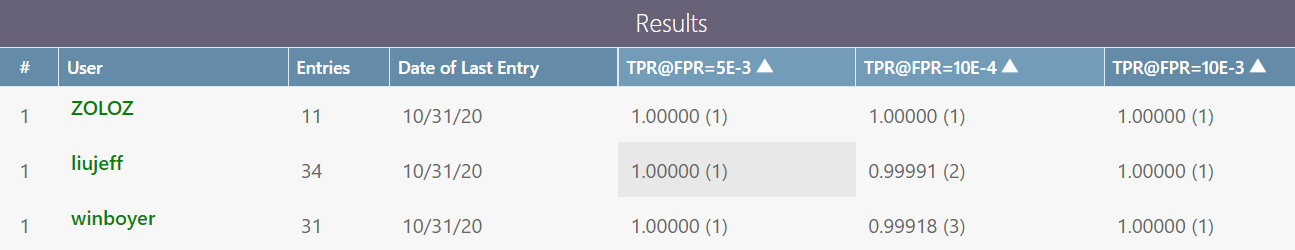}
    \caption{Final results for the 2020 CelebA-Spoof Face Anti-Spoofing Challenge.}
    \label{fig:celeb-results}
\end{figure*}

The Face Forgery Analysis Challenge 2021\footnote{\url{https://competitions.codalab.org/competitions/33386}} is a competition hosted at the 2021 IEEE/CVF Conference on Computer Vision and Pattern Recognition (CVPR 2021). It is organised by researchers from a number of organisations in China including universities and SenseTime Research (the research arm of SenseTime\footnote{\url{https://www.sensetime.com/}}, one of the major AI ``unicorns'' in China). The challenge aims to advance the state-of-the-art in detection of photo-realistic manipulation of images and videos. Participants are able to use a large annotated face dataset (i.e., the ForgeryNet dataset) that was obtained by applying a number of techniques for manipulation (15) and perturbation (36) 
to train their solutions. The phases comprise of Forgery Image Analysis, Forgery Video Analysis, Forgery Video Temporal Localization phases, and the final phase (i.e., ``private test'') where participants' models will be tested against an unseen dataset. The following metrics will be used~\cite{He_2021_CVPR}: AUC, average precision (AP) at some ``temporal Intersection over Union'' (AP@tIoU) compared to a threshold $t\in[0.5,0.95]$, and average recall (AR) at $K$ (AR@$K$) where $K$ is the top $K$ labels returned for multi-class classifiers.

The 2020 CelebA-Spoof Face Anti-Spoofing Challenge\footnote{\url{https://competitions.codalab.org/competitions/26210}} was hosted at the $16^{\textit{th}}$ European Conference on Computer Vision (ECCV 2020). The challenge ran between August and October 2020, and aimed to advance the state-of-the-art in detecting ``\textit{whether a presented face is live or spoof}''~\cite{ZYSLYXXXLLLCGLLGHHLCQZLJH2021}. The organisers made the face CelebA-Spoof dataset available for the competition containing 
rich annotation across a range of attributes. The competition only had one phase where participants submitted their solutions to be evaluated against a test dataset; the spoof class was considered as ``positive'' and the live class as ``negative''. Metric TPR@FPR was used and collected at three points where the TPR when $\text{FPR}=10^{-4}$ determined the final ranking. The top three finalists (see Figure~\ref{fig:celeb-results}) used deep learning models ResNet, EfficientNet-B7, and a novel architecture combining Central Difference Convolutional Networks (CDCN) and Dual Attention Network (DAN). The two top ranked solutions used different strategies to boost their models' performance: a heuristic voting scheme was used by the top-ranked solution, and a weight-after-sorting strategy was used by the second ranked solution.

The 2021 CSIG Challenge\footnote{\url{http://challenge.csig.org.cn/}} is the second edition of a challenge organised by the China Society of Image and Graphics\footnote{\url{http://www.csig.org.cn/}}. The 2021 challenge has the Fake Media Forensic Challenge\footnote{\url{http://fmfcc.net/}} as its $6^{\textit{th}}$ track, co-organised by CSIG's  Digital Media Forensics and Security Technical Committee\footnote{\url{http://www.csig.org.cn/detail/2450}} and Institute of Information Engineering, Chinese Academy of Sciences\footnote{\url{http://www.iie.ac.cn/}}. This track has two tasks, one on deepfake video detection, and the other on deepfake audio/speech detection. For the deepfake video detection task, the dataset used contains a public training set with 10,000 sound-free face videos (including 4,000 fake videos), a public test set with 20,000 face videos (the percentage of deepfake videos is unknown to participants), and a private test set that will be determined and used at the final session for selecting the winners. All videos contain faces of Eastern Asian people, and cover a wide range of parameters such as multiple resolutions and encoding quality factors, the use of blurring or sharpening filters, and added noise. Deepfake videos were created using public tools including DeepFaceLab~\cite{perov2020}, 
Faceswap\footnote{\url{https://github.com/deepfakes/faceswap}}, Faceswap-GAN, Recycle-GAN~\cite{bansal2018} and ALAE (Adversarial Latent Autoencoders)~\cite{Pidhorskyi_2020_CVPR}. For the deepfake audio/speech detection task, the dataset used contains a public training set with 10,000 speech samples (including 6,000 fake ones), a public test set with 20,000 face videos (the percentage of deepfake videos is unknown to participants), and a private test set for the final session (the same as the deepfake video detection task). The tools used for generating the fake speech samples include TTS (text-to-speech) voice synthesis tools and VC (voice conversion) tools. The main TTS tools used include open-source tools such as DeepVoice, TensorFlowTTS\footnote{\url{https://github.com/TensorSpeech/TensorFlowTTS}} and GAN-TTS~\cite{binkowski2019} and commercial software tools such as those from iFlytek\footnote{\url{https://en.wikipedia.org/wiki/IFlytek}} and IBM. The main VC tools used include Adaptive-VC and CycleGAN-VC~\cite{kaneko2017}. For both deepfake detection tasks, the performance metric used is log loss.

2020 China Artificial Intelligence\footnote{\url{https://ai.xm.gov.cn/}} was the second edition of a Chinese AI competition open for the general public to participate, organised by the municipal government of the City of Xiamen in China. In 2020, it had two sub-competitions, Multimedia Information Recognition Technology Competition\footnote{\url{https://ai.xm.gov.cn/competition/competition-detail.html?id=2200075d26e840b1b9ffd10633d6a9bf}} and Language and Knowledge Technology Competition\footnote{\url{https://ai.xm.gov.cn/competition/competition-detail.html?id=0000075d26e840b1b9ffd10633d6a9bf}}. The Multimedia Information Recognition Technology Competition included two tasks on deepfakes: one on deepfake video detection\footnote{\url{https://ai.xm.gov.cn/competition/project-detail.html?id=2110df4351414ffe8eae1df3c3507e95&competeId=2200075d26e840b1b9ffd10633d6a9bf}} and one on deepfake audio/speech detection\footnote{\url{https://ai.xm.gov.cn/competition/project-detail.html?id=89cd6fad92a346009c4b5a6690828da7&competeId=2200075d26e840b1b9ffd10633d6a9bf}}. The deepfake video detection task used 3,000 videos, and log loss was used as the sole performance metric. The deepfake audio/speech detection task used 20,000 audio samples (mostly in Chinese, and the remaining in English), and EER was used as the sole performance metric. For both tasks, the ratio between real and deepfake samples was 1:1. We did not find where to download the datasets used for the tasks nor a more detailed technical description of the datasets. For the deepfake video detection tasks, the top two winning teams (with an A prize) were from Netease (Hangzhou) Network Co., Ltd. and Beijing RealAI Technology Co., Ltd., followed by three other teams winning a B prize: Xiamen Fuyun Information Technology Co., Ltd.; Institute of Computing Technology, Chinese Academy of Sciences; and Wuhan Daqian Information Technology Co., Ltd. For the deepfake audio/speech task, there was no team winning an A prize, but one team winning a B prize: SpeakIn Technologies Co., Ltd. The final results of some teams were published, but some teams were allowed to hide their results. We did not find a detailed technical report summarising the results and explaining the work of the winning teams.

One of the B-prize winning team is from Beijing RealAI Technology Co., Ltd., a Chinese company active in deepfake-related R\&D.

\subsection{Generation of Manipulated Media\label{Subsec:challenges_deepfake_generation}}

\begin{figure*}[!htb]
\centering
\includegraphics[width=0.55\textwidth]{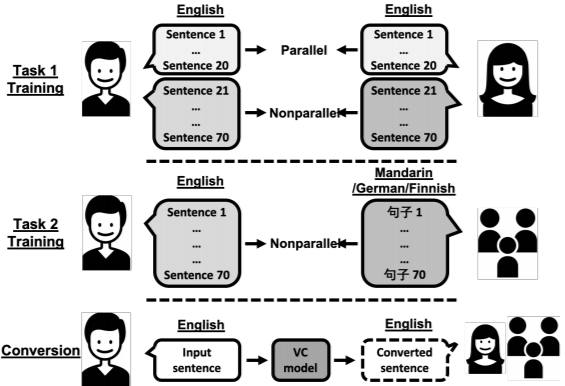}
\caption{Illustration of tasks for the Voice Conversion Challenge 2020, extracted from~\cite{YHTYDKLT2020}.}
\label{fig:VCC}
\end{figure*}

The Voice Conversion Challenge\footnote{\url{http://www.vc-challenge.org/}} is a biennial competition that has been running since 2016. The challenge and the corresponding workshop, hosted at the INTERSPEECH conference\footnote{\url{https://www.isca-speech.org/iscaweb/index.php/conferences/interspeech}}, is supported by the SynSig (Speech Synthesis Special Interest Group)\footnote{\url{https://www.isca-speech.org/iscaweb/index.php/sigs?layout=edit&id=117}} of the International Speech Communication Association (ISCA)\footnote{\url{https://www.isca-speech.org/}}. Its aim is to promote progress in voice conversion (VC) technology that can be applied to a number of positive and negative use cases, such as spoofing voice biometric systems. The 2020 challenge focused on speaker conversion, a sub-problem of VC, and included two tasks. For the first task ``intra-lingual semi-parallel voice conversion'', participants had to develop 16 VC systems (speaker-pair combinations) including male and female speakers and English sentences, using the provided Voice Conversion Challenge 2020 database v1.0 for training (refer to Section~\ref{sec:datasets}). For the second task ``cross-lingual voice conversion'', participants had to develop 24 VC systems, also including male and female speakers, but uttering sentences in three languages (Finnish, German and Mandarin), based on the provided training dataset. Figure~\ref{fig:VCC} illustrates the process of training and generation of VC systems. 

Submissions were evaluated for ``\textit{perceived naturalness and similarity through listening tests}''\footnote{\url{http://www.vc-challenge.org/}}. As such, the organisers used \emph{subjective evaluation}~\cite{YHTYDKLT2020} and recruited both native and non-native English speakers (i.e., Japanese native speakers) via crowd-sourcing for the listening tests. Naturalness (answering the question ``\textit{How natural does the converted voice sound?}'') was measured using the metric MOS (covered in  Section~\ref{Subsec:datasets_subjective_quality}), and similarity (answering the question ``\textit{how similar the converted voice sound comparing source and target speakers?}'') was measured in terms of speaker recognition as ``same'' or ``different'', as elaborated by~\citet{WWY2016}. Tests also focused on the effects of language differences on the performance of VC systems submitted to the competition. The most popular CNN/RNN/GAN-based VC systems submitted used WaveNet, WaveRNN, and Parallel WaveGAN. Results indicated that, in terms of similarity, the best performing VC systems were as good as natural speech but none reached human-level naturalness for task 1; scores were lower for task 2 which was more complex~\cite{YHTYDKLT2020}. The organisers of the 2020 competition also used \emph{objective evaluation}~\cite{DKHLYYTT2020}. The metrics used for evaluation of speaker similarity were: equal error rate (EER), false acceptance rate of target ($P^{\text{tar}}_{\text{fa}}$), miss rate of source ($P^{\text{src}}_{\text{miss}}$), and cosine similarity of speaker embedding vectors (cos-sim) according to Eq.~\eqref{eq:COS} where $A$ is the speaker embedding vectors for the converter audio and $B$ is the speaker embedding vectors for the original audio. The performance of the VC systems as a spoof countermeasure was also evaluated using EER, while to evaluate the quality of the subjective MOS obtained via listening tests, a DL-based model to predict MOS, called MOSNet~\cite{LFHWYTW2021}, was used. Lastly, to evaluate intelligibility of the converted transcribed speech, in comparison with the original transcribed speech, the word error rate (WER)~\cite{AR2018} was used. WER is calculated according to Eq.~\eqref{eq:WER} where $I$ refers to insertions, $D$ refers to deletions, $S$ refers to substitutions, and $N$ refers to the total number of words in the original transcript.

\begin{align}
\text{cos-sim}(A,B) & = \frac{A \times B}{\parallel A \parallel \parallel B \parallel}\label{eq:COS}
\end{align}

\begin{align}
\text{WER} & = \frac{I + D + S}{N} \times 100\label{eq:WER}
\end{align}

The Deepfake Africa Challenge (2021)\footnote{\url{https://zindi.africa/competitions/deepfake-africa-challenge}} is a new initiative of the AI Africa Expo, in partnership with a film and media production company (Wesgro) and the African Data Science competition platform Zindi. Its aim is ``\textit{to create convincing deepfakes to highlight the power of this synthetic media, illustrating its creative potential for exploitation for both positive and negative outcomes and focusing debate about its ethical use / mis-use in an African context}''. Eligible participants were required to be citizens and residents of the African continent. Submissions, accepted up to end of July 2021, can be either video or audio. Evaluation of submissions is defined in terms of artistic creativity, relevance of challenge topic, and innovation in the process of generation as long as participants use tools and packages publicly available. The top three finalists will receive a prize, present their work at the Expo, and will have to grant copyrights to Zindi. Unlike the other competitions reviewed in this section, which were focused on advancing the state-of-the-art in detection of synthetic or manipulated media, this competition focused on the generation of deepfake which seems more humanities-centred. This is a trend observed in arts~\cite{digitnews2021} and culture~\cite{medium2021}.

\subsection{Generation and Detection of Manipulated Media} 

The DeepFake Game Competition (DFGC)\footnote{\url{https://competitions.codalab.org/competitions/29583}} is in its first edition, hosted at the 2021 International Joint Conference on Biometrics (IJCB 2021). Its organisers are mainly from the Institute of Automation Chinese Academy of Sciences (CASIA). The idea of the competition was to promote an adversarial game between agents pushing for advances in both deepfake creation and detection. In order to achieve this, a 6-stage protocol was designed interleaving three creation phase (C-phase) and detection phase (D-phase), typically one week apart; submissions closed in April 2021. Both C-phases and D-phases were bound to the Celeb-DF (v2) dataset~\cite{LYSQL2020}, containing 6,229 videos (590 real/original videos and 5,639 fake/manipulated videos), for training purposes. As such, submissions to a C-phase would consist of datasets extracted from Celeb-DF (v2) which included novel face-swap approaches to obtain evaluation results. Submissions to a D-phase would consist of detection models/codes to obtain evaluation results. The models submitted for a D-phase were evaluated against the datasets submitted for the previous C-phase~\cite{PFWDLLLSCCH2021}. The metrics used for evaluation\footnote{\url{https://competitions.codalab.org/competitions/29583\#learn_the_details-evaluation}} were: a detection score, used for evaluation of a D-phase, and a creation score, used for evaluation of a C-phase. The top three finalists for the detection phase employed CNN-based classifiers EfficientNet-B3, Efficientnet-B0 and EfficientNetV2.

The Detection Score ($D_{S}$) metric 
captures the models' ability to correctly classify fake images submitted to the previous C-phase against a set of real images in the CelebDF test dataset. It is calculated using 
Eq.~\eqref{eq:DS}, where $N_{C}$ is the number of valid submissions of created synthesis test sets in the last C-phase. 
\begin{align}
D_{S} & = \sum_{i=1}^{N_{C}} \frac{\text{AUC}_{i}}{N_{C}} \label{eq:DS}
\end{align}

The Creation Score ($C_{S}$) metric used to evaluate creation models submitted to this challenge 
is calculated by Eq.~\eqref{eq:CS}, where $N_{D}$ is the number of valid submissions of detection methods in the last D-phase, the noise score ($S^{\text{noise}}$) penalises noisy images, the other three parts of the equation relate to the following\footnote{\url{https://competitions.codalab.org/competitions/29583\#learn_the_details-evaluation}}: ``\textit{ID level similarity to the donor ID, image level similarity to the target frame, and the deception ability against detection models. ID level similarity is scored by a face recognition model using dot product of two ID features (fake face ID and donor ID). The image level similarity is scored by SSIM [Structural Similarity Index] to make sure the face-swapped image is similar to the corresponding target image in content and quality}''. 

\begin{align}
C_{S} & = S^{\text{noise}}(I_{\text{fake}}) + B + C + D \label{eq:CS} \\
B & = S^{\text{ID}}(\text{ID}_{\text{fake}},\text{ID}_{\text{donor}}) \nonumber \\
C & = S^{\text{SSIM}}(I_{\text{fake}},I_{\text{target}}) \nonumber \\
D & = 2 \times \sum_{i=1}^{N_{D}} \frac{1-\text{AUC}_{i}}{N_{D}}  \nonumber
\end{align}

%\subsection{Discussion: Challenges, Competitions and Benchmarks}

\citet{PFWDLLLSCCH2021} observed    
a commonality between the three winning teams for the creation task, i.e., the use of the FaceShifter~\cite{LBYCW2020} framework for face swapping. They highlighted two overall reflections about the competition: (1) the limited diversity of the deepfake datasets submitted and the use of repetitive methods to generate them, and (2) the limited size of the Celeb-DF (v2) dataset itself flagging the need for a larger dataset for next year's competition. The organisers of the competition also applied the top two detection models to unseen datasets (DFDC and FaceForensics++) and noticed that they do not generalise well.  

%%%%%%%
\section{A Meta-Review of Deepfake-Related Surveys \label{sec:meta-review}}

This section presents a meta-review of 12 selected deepfake-related survey papers, including eight published in English \cite{TWPW2020, DW2021, Lyu2020, YH2020, TVFMO2020, ZDZD2020, Verdoliva2020, ML2021} and four published in Chinese \cite{LLZCXZHHZLC2020, LJWLDCYK2021, TFYWW2020, BLD2020}. 
It covers the following aspects in a systematic manner: definitions and scope, performance metrics, datasets, challenges/competitions/benchmarks, performance comparison, key challenges and recommendations. 

The meta-review 
aims at drawing some high-level insights for monitoring future development of deepfake-related technologies and their applications.

\subsection{Definitions and Scope}

As we discussed in Section~\ref{sec:definitions}, among researchers, practitioners and law makers there is no  universally accepted definition of ``deepfake'' as a term. This is also reflected in how the authors of the 12 survey papers considered this aspect. Most authors talked about the history of deepfakes and pointed out that the term reflects the combination of ``deep learning'' and ``fake'', but some used a broader definition, e.g., \citet{Lyu2020} defined deepfake as ``\textit{high quality fake videos and audios generated by AI algorithms}''. Some authors also referred to deepfake-related legislations, but none of them pointed out that the definitions in some such legislations are completely different from the more technical definitions involving the use of deep learning. No authors discussed the blurred boundary between deepfakes and non-deepfakes, although some surveys actually cover both, e.g., \citet{TFYWW2020} focused on speech forgery and did not explicitly highlight ``deepfake''.

In terms of the scope, while some authors (correctly) considered all types of media that can be produced by deepfake-related techniques \cite{TWPW2020, Lyu2020, LLZCXZHHZLC2020, LJWLDCYK2021}, some considered only a narrow scope, e.g., authors of \cite{TVFMO2020, ZDZD2020, YH2020, BLD2020} considered only videos, and only authors of \cite{Verdoliva2020, DW2021} have considered images and videos. Another phenomenon we observed is that many authors focused more on face images and videos, and authors of three surveys \cite{DW2021, YH2020, TVFMO2020} even limited the definition of ``deepfake'' to such a narrow scope:
\begin{itemize}
\item \citet{DW2021} defined it as ``\textit{a technology which creates fake images or videos of targeted humans by swapping their faces [by] another character saying or doing things that are not absolutely done by them and humans start believing in such fake as it is not always recognisable with the everyday human eye}'';  
\item \citet{YH2020} considered deepfake as a technique allowing ``\textit{any computer user to exchange the face of one person with another digitally in any video}''; and 
\item \citet{TVFMO2020} defined it as ``\textit{a deep learning based technique able to create fake videos by swapping the face of a person by the face of another person}''. 
\end{itemize}
Such unnecessarily narrow definitions and scopes can lead to confusion and do not help exchanges between researchers and practitioners working on different types of deepfakes.

\begin{figure}[!h]
\centering
\includegraphics[width=\linewidth]{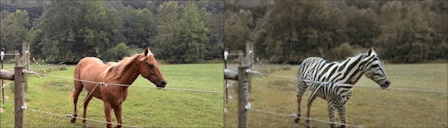}
\caption{An image of a horse (left) and a deepfake image generated using the image2image technique proposed in~\cite{zhou2017} (right).}
\label{fig:horse2zebra}
\end{figure}

We call on more researchers to accept a broader definition of ``deepfake'' so that highly realistic/natural media of any kind generated by a sophisticated automated method (often AI-based) is considered deepfake. Here, we provide two examples of such a broader definition: the image2image (or pixel2pixel) technique~\cite{zhu2017} that allows the production of deepfake images and videos of any objects (e.g., the ``horse2zebra'' deepfake image shown in Figure~\ref{fig:horse2zebra}), and the the so-called ``deepfake geography~\cite{zhao2021}'', where AI-based techniques are used to generate realistic-looking satellite images.

Another important fact missed or not sufficiently discussed by authors of all the 12 surveys is that deepfake techniques can be used for positive applications, e.g., creative arts, entertainment and protecting online users' privacy. We call for more researchers and practitioners to follow the proposal in the 2020 Tencent AI White Paper~\cite{tencent2020} to start using the more neutral-sounding term ``deep synthesis''. Accordingly, we can use different words for different types of data generated using ``deep synthesis'' techniques, e.g., ``deep art'', ``deep animation'', ``deep music'', and ``deepfake''. While authors of the 12 survey papers did not recognise the positive applications of ``deepfake'' technologies, some other researchers did, e.g., organisers of the Voice Conversion Challenge 2020\footnote{\url{http://www.vc-challenge.org/}} who said the VC technology (for speech deepfake) ``\textit{is useful in many applications, such as customizing audio book and avatar voices, dubbing, movie industry, teleconferencing, singing voice modification, voice restoration after surgery, and cloning of voices of historical persons}''.

\subsection{Performance Metrics}

Surprisingly, none of the 12 surveys have covered performance metrics explicitly. Some directly used performance metrics to explain and compare performance of covered deepfake generation and detection methods. The most used performance metrics include accuracy, ERR, and AUC. This may be explained by the page constraints of such survey papers, which did not allow the authors to extend their coverage significantly to cover performance metrics systematically. The subjective quality of deepfakes is an area least covered by the surveys, which seems related to an unbalanced coverage on deepfake generation and deepfake detection in terms of performance evaluation and comparison (the former much less than the latter).

\subsection{Datasets}

Many of the 12 survey papers list a number of deepfake-related datasets, but none of them have coverage as complete as ours shown in Section~\ref{sec:datasets}. For instance, none of the surveys have covered the Voice Conversion Challenge 2016/2018/2020 datasets and the ASVspoof 2019/2021 datasets are covered briefly only in two surveys \cite{LJWLDCYK2021, TFYWW2020}. In addition, more recent deepfake datasets especially those released in 2021 are also not covered by any of the surveys. We believe that our Section~\ref{sec:datasets} is the most comprehensive review of deepfake-related datasets so far.

Some survey papers include datasets that are likely deepfakes, e.g., \citet{Verdoliva2020} covered many general fake image datasets where the manipulated images were not generated by deep learning or even AI-based methods, and some surveys (e.g., \cite{LJWLDCYK2021}) mentioned ASVspoof 2015 datasets but we did not see the use of deep learning for generating data used in the dataset.

\subsection{Challenges, Competitions and Benchmarks}

Many surveys cover deepfake-related challenges, competitions and benchmarks. The coverage is, however, mostly limited, and some challenges (e.g., the Voice Conversion Challenge 2016/2018/2020 and the two Chinese challenges we covered in Section~\ref{Sec:challenges}) are not covered by any of the surveys. The level of detail of challenges, competitions and benchmarks is also normally limited, compared with what we chose to include in Section~\ref{Sec:challenges}. Similar to the datasets we covered in Section~\ref{sec:datasets}, we believe that our coverage of deepfake-related challenges, competitions and benchmarks in Section~\ref{Sec:challenges} is also the most comprehensive so far.

\subsection{Performance Comparison}

Most surveys have a good coverage of related methods for deepfake generation and detection, but only some explicitly covered performance comparison between different methods \cite{TVFMO2020, ML2021, LJWLDCYK2021}.

Among all the survey papers, \citet{LJWLDCYK2021} conducted the most comprehensive study on performance of different deepfake detection methods. In addition to showing the performance metrics of a number of deepfake detection methods in Table~3 of \cite{LJWLDCYK2021}, they also looked at general characteristics and issues of different types of deepfake detection methods, as shown in Table~\ref{tab:LJWLDCYK2021_comparison}. Furthermore, they also looked at research on robustness of deepfake detection methods against adversarial samples, referring to some work that showed a lack of such robustness. 
\begin{table*}[!tb]
\caption{Comparison of different deepfake detection methods as shown in Table~4 of \cite{LJWLDCYK2021}.}
\label{tab:LJWLDCYK2021_comparison}
\begin{tabularx}{\linewidth}{lXX}
\toprule
\textbf{Method} & \textbf{Characteristics} & \textbf{Issues}\\
\midrule
Image forensics based & More mature, more explainable & Image-only, robustness against lossy compression\\
Biological signals based & Specific signals, local information & High error rate for lossily compressed videos, some features unavailable, less accurate\\
Image forgery detection based & Local information, effective for low-quality deepfakes & Less generalisable, less accurate\\
GAN-fingerprinting based & GAN-specific & Data and algorithm dependency, less generalisable\\
Data-driven & Big data, rich information, high accuracy & Data dependency, sensitive to unknown data and lossy compression\\
\bottomrule
\end{tabularx}
\end{table*}

Due to quality issues of many deepfake-related datasets (discussed in Section~\ref{Subsec:datasets_subjective_quality}), we need to treat any performance metrics and comparison of different detection methods with caution. Without testing all methods on a sufficiently large, diverse and high-quality deepfake dataset, the performance comparison results can be misleading. This highlights the importance of having more challenges, competitions and benchmarks to encourage performance comparison on standard datasets and using consistent performance metrics.

\subsection{Challenges and Recommendations}

The authors of some surveys identified some key challenges and future research directions for the deepfake community.

Not surprisingly, how to develop more robust, scalable, generalisable and explainable deepfake detection methods is one of the most discussed key challenges and also a major future research direction \cite{TWPW2020, DW2021, Lyu2020, YH2020, Verdoliva2020, LLZCXZHHZLC2020, LJWLDCYK2021, TFYWW2020, BLD2020}. Considering the arms race between deepfake generation and detection, this research direction will likely remain the hottest topic in deepfake research.

A couple of surveys \cite{Verdoliva2020, LJWLDCYK2021} mentioned fusion as a key future research direction, where ``fusion'' refers to combining different methods (e.g., combining multiple detectors of different types) and data sources (e.g., jointly considering audio-visual analysis) to achieve better performance for deepfake detection. \citet{Lyu2020} suggested that, for detection of deepfake videos, we need to consider video-level detection more, which can be considered fusion of detection results of all video frames.

The authors of three surveys, \citet{Lyu2020} , \citet{DW2021} and \citet{YH2020}, argued that better (higher-quality, more up-to-date, and more standard) deepfake datasets are needed to develop more effective deepfake detection methods. \citet{Lyu2020} also suggested that we need to consider \emph{social media laundering} effects in training data and improve the evaluation of datasets. We agree with them on these points.

\citet{TFYWW2020} suggested that low-cost deepfake generation/detection should be considered as a future research direction. This is a valid recommendation since lightweight methods will allow less powerful computing devices (e.g., IoT devices) to benefit from such technologies.

Two Chinese surveys \cite{LLZCXZHHZLC2020, LJWLDCYK2021} also mentioned the need to have new deepfake-related legislations combating malicious use of deepfakes and the need to train end users such as journalists. This is likely an area where  interdisciplinary research can grow.

There are also other ad-hoc recommendations given by the authors of some surveys. For example, \citet{Lyu2020} argued that deepfake detection should be considered a (more complicated) multi-class, multi-label and local detection problem. \citet{TVFMO2020} discussed specific research directions for different deepfake generation methods (face synthesis, identity swap, attribute manipulation, and expression swap). \citet{LLZCXZHHZLC2020} and \citet{LJWLDCYK2021} recommended more active defence mechanisms such as using digital watermarking and blockchain technologies to build trustworthy media frameworks against deepfakes.

\section{Conclusion\label{sec:conclusion}}

The rapid growth in the capability to manipulate media or create synthetic media which look realistic and natural paved the way for deepfakes. At first, this paper adopted a critical approach to look at different definitions of the term ``deepfake''. In that regard, we point out the different contradicting definitions and call for the wider community to consider how to define a new term that has a more consistent scope and meaning. For instance, replacing ``deepfake'' by ``deep synthesis'' can be more inclusive by embracing positive applications of deepfake techniques, e.g., in entertainment and for simulation purposes.

This paper provided a comprehensive overview of multiple aspects of the deepfake ecosystem drawing from the research literature and other online sources published in two languages: English and Chinese. It covers commonly used performance metrics and standards, related datasets, challenges, competitions and benchmarks. It also presents a meta-review of 12 selected deepfake-related survey papers published in 2020 and 2021, covering not only the above mentioned aspects, but also highlighting key challenges and recommendations.

\bibliographystyle{ACM-Reference-Format}
\bibliography{deepfake}

\end{document}